\pdfoutput=1
\documentclass[journal,twoside]{IEEEtran}
\usepackage{cite}
\usepackage{amsmath,amssymb,amsfonts}
\usepackage{graphicx}
\usepackage{textcomp}
\usepackage{framed,multirow}

\usepackage{amssymb}
\usepackage{latexsym}
\usepackage{threeparttable}

\usepackage{url}

\usepackage[table]{xcolor}
\usepackage{caption,subcaption,booktabs}
\usepackage{adjustbox}

\usepackage{amsmath}
\usepackage{color}
\usepackage{hyperref}
\hypersetup{
	colorlinks,
	linkcolor=[rgb]{0.0, 0.8, 0.2},
	citecolor=[rgb]{0.0, 0.8, 0.2},
	urlcolor =[rgb]{0.0, 0.8, 0.2}
}

\usepackage{rotating}

\usepackage{hyperref}

\usepackage{amsmath}
\usepackage{bm}

\definecolor{newcolor}{rgb}{.8,.349,.1}

\newif\ifshowchanges
\showchangesfalse
\newcommand{\trinh}[1]{\ifshowchanges\textcolor{blue}{#1}\else#1\fi}
\newcommand{\revisioncolor}{\ifshowchanges\color{blue}\fi}

\newif\ifshowchangesrtwo
\showchangesrtwofalse

\def\BibTeX{{\rm B\kern-.05em{\sc i\kern-.025em b}\kern-.08em
    T\kern-.1667em\lower.7ex\hbox{E}\kern-.125emX}}
\begin{document}
\title{MiTHras: Task-specific Hierarchical Semi-supervised Contrastive Masked Autoencoder for Mitotic Figure Analysis}
\author{Trinh T.L. Vuong,  Simon Graham,  Quoc Dang Vu, Phat T.H. Ho, Jeewoo Lim, Mostafa Jahanifar,  Nasir Rajpoot\textsuperscript{*}, Jin T. Kwak\textsuperscript{*}, \IEEEmembership{Member, IEEE}
\thanks{This work was supported by the National Research Foundation of Korea (NRF) grant (No. RS-2025-00558322), Korea Institute for Advancement of Technology (KIAT) grant (No. P0022543), and the Innovate UK grant (No. 10040491). (Corresponding authors:  Nasir Rajpoot and Jin T. Kwak.)}
\thanks{Trinh T.L. Vuong is with Department of Pathology, Brigham and Women’s Hospital, Harvard Medical School, Boston, USA (e-mail: tlvuong@bwh.harvard.edu).}
\thanks{Jeewoo Lim and Jin T. Kwak are with School of Electrical Engineering, Korea University, Seoul 02841, Korea (e-mail: \{jeewoolim and jkwak\}@korea.ac.kr).}
\thanks{Phat T.H. Ho is with the Department of Pathology, University Medical Center Ho Chi Minh City, Ho Chi Minh City, Vietnam (e-mail: phat.hth@umc.edu.vn).}
\thanks{Simon Graham, Quoc Dang Vu, Mostafa Jahanifar, and Nasir Rajpoot are with Histofy Ltd, Coventry, United Kingdom (e-mail: \{s.graham, qd.vu, m.jahanifar, n.rajpoot\}@histofy.ai).}
}

\maketitle

\begin{abstract}
\trinh{Mitotic figure (MF) analysis supports tumor grading and prognostic assessment, but automated models remain sensitive to differences in tissue type and image acquisition. We present MiTHras, a task-specific pretraining framework that combines pseudo-label-guided image- and token-level contrastive learning with masked reconstruction. We construct TCGA-MF-Pseudo, a corpus of 1.8 million cell-centered images from 14 TCGA cohorts spanning 11 organ sites. Comprehensive evaluation on MF classification, detection, count-based survival prediction, and subtype classification demonstrates the efficacy of MiTHras. It achieves the highest mean F$_1$ on all three MF classification benchmarks and both subtype benchmarks. MiTHras also outperforms general-purpose and pathology foundation encoders by a larger margin under frozen-encoder linear probing than under full fine-tuning. Although detection gains are modest due to a shared candidate-detection stage, ablations confirm that token-level supervision improves typical-versus-atypical classification and linear probing. These findings establish that MiTHras yields robust, transferable representations for automated mitotic activity assessment.
}
\end{abstract}


\section{Introduction} \label{section:intro}

\trinh{The mitotic count is a marker of cell proliferation used in cancer diagnosis, tumor grading, and prognostic assessment \cite{von2013prognostic}. Pathologists determine it by counting mitotic figures (MFs) within a specified tissue area, following tumor-specific grading protocols \cite{jung2022update, who2022bluebooks}. It is a core component of the Nottingham grading system for invasive breast carcinoma \cite{elston1991pathological} and FNCLCC grading in soft-tissue sarcoma \cite{coindre2006grading}, drives risk stratification in gastrointestinal stromal tumors \cite{miettinen2006gist}, and defines tumor grade together with Ki-67 in gastroenteropancreatic neuroendocrine tumors \cite{rindi2018nen}. Automated MF analysis provides a total count with precise locations, allowing pathologists to inspect the underlying detections. These outputs complement weakly supervised models that predict slide-level clinical endpoints. We consider two metrics for mitotic activity. The \emph{mitotic hotspot count} reports the highest MF count within a fixed area. The \emph{mitotic density} denotes the total MF count per unit tissue area. A localized active focus can therefore produce a high hotspot count but a low overall density.}

With the increasing availability of digitized whole-slide images (WSIs), a growing number of datasets have been developed for mitotic figure detection across diverse cancers and organs \cite{aubreville2023comprehensive}. However, these datasets remain limited in size and diversity due to the high cost of annotation, restricting their ability to support large-scale model training. Consequently, \textit{domain shift} remains a major challenge, leading to degraded performance when models are applied to unseen cancer types or data sources \cite{aubreville2024domain}.
In computational pathology, pretraining plays a key role in improving model generalization under limited annotations. Self-supervised learning (SSL) has emerged as an effective strategy by leveraging large-scale unlabeled histopathology data, through contrastive learning (e.g.\ SimCLR \cite{chen2020simple}), self-distillation, and \trinh{masked image modeling (MIM)} (e.g.\ MAE \cite{he2022masked}). These methods have been adapted to pathology to train general-purpose feature extractors such as CTransPath-MoCov3 \cite{wang2022transformer} and Virchow \cite{vorontsov2024foundation}. While these models generalize strongly across tasks, they are designed as generic feature extractors rather than optimized for a specific task such as MF analysis.

Most pretrained models are used as frozen feature extractors with task-specific heads for downstream applications such as WSI classification and survival prediction \cite{vorontsov2024foundation, song2024multimodal}. Further adaptation can improve performance \cite{lee2024benchmarking, nguyen2024camp}, for example through knowledge distillation \cite{le2025moma}. However, discrepancies between pretraining objectives and downstream tasks introduce a domain gap that limits direct transfer.
Moreover, most pretrained (foundation) models are trained on relatively large image patches (e.g., 112$\times$112 $\mu m$), whereas MFs are typically assessed at much finer scales (e.g., 32$\times$32 $\mu m$) \cite{jahanifar2024mitosis}. This mismatch in resolution and context limits their effectiveness, as does the presence of numerous visually similar mimickers such as apoptotic cells and lymphocytes, which calls for task-specific rather than generic features. Recent pathology foundation models generalize well, but their size makes them expensive for dense inference tasks such as mitotic detection. With limited and imbalanced datasets, these challenges motivate efficient, task-specific pretraining.

To address these challenges, we propose MiTHras, a \textbf{Mi}totic figure analysis model based on \textbf{T}ask-specific \textbf{H}ierarchical semi-supervised cont\textbf{ras}tive learning with a masked autoencoder. 
\trinh{Through a single, unified objective, it integrates image-label contrast, which aligns the representations of crops that share a mitotic or mimicker pseudo-label, token-level contrast, which aligns local features using pseudo bounding boxes, and masked reconstruction that preserves image information. We evaluate MiTHras on MF classification, detection, count-based survival prediction, and subtype classification.
}

In summary, our contributions are as follows:
\begin{itemize}
    \item \textbf{MiTHras Framework:} \trinh{We introduce a task-specific hierarchical semi-supervised framework integrating pseudo-label-guided image- and token-level supervision with masked reconstruction. 
    }
    \item \textbf{Large-scale Dataset:} We introduce TCGA-MF-Pseudo, a pseudo-labeled corpus that substantially broadens the diversity and scale of training data available for MF analysis.
    \item \textbf{Comprehensive Evaluation:} \trinh{We evaluate classification, detection, count-based survival prediction, and subtype classification against MF models, pretraining methods, and pathology foundation models under fine-tuning and linear probing settings.
    }
\end{itemize}

\section{Related work}

\subsection{Mitotic figure detection} 
Deep learning approaches for MF analysis include classification, object detection, segmentation, and two-stage pipelines. Methods often incorporate domain adaptation to address \textit{domain shift} caused by variations in scanners, staining, and tissue types.
Object detection-based methods have shown effectiveness in localizing MFs; \cite{wilm2021domain}, for instance, introduced a domain-adversarial RetinaNet to reduce domain shift. Segmentation-based methods instead leverage pixel-level predictions to capture fine-grained mitotic features: MaskMitosis \cite{sebai2020maskmitosis} adopts Mask R-CNN with a ResNet-FPN backbone \cite{he2016deep}, and other works combine segmentation with Mask R-CNN for detection \cite{fick2021domain}.

Two-stage models typically detect candidate regions and subsequently classify them. FoCasNet \cite{wang2024novel} used a feature pyramid network for detection and a ResNet \cite{he2016deep} for classification, and \cite{fick2021domain} combined Mask R-CNN for segmentation with an ensemble of DenseNet and ResNet. MDFS \cite{jahanifar2024mitosis} adopted Efficient-UNet-b0 for segmentation and EfficientNet for classification, achieving top performance in both \trinh{MIDOG21 \cite{aubreville2023comprehensive} and MIDOG22 \cite{aubreville2024domain}} challenges. More recently, vision-language models have been applied to MF analysis: \cite{ding2024improving} paired MF images with descriptive captions to support captioning and visual question answering, and by integrating modality and scanner information demonstrated improved cross-domain adaptability.

Despite these advancements, most existing methods heavily rely on supervised data across multiple domains and require domain alignment strategies. In this work, we propose a pretrained framework designed to learn from diverse cancer types and scanner sources, aiming to enhance the generalization ability in MF detection.

\subsection{\trinh{Representation learning in computational pathology}}

\trinh{Hierarchical pathology models combine information across spatial scales. HIPT \cite{chen2022hipt} stacks self-supervised transformers from patch to region to slide, CTransPath \cite{wang2022transformer} captures multiscale features via windowed attention. Attention-based multiple-instance learning aggregates patch features for slide-level prediction \cite{ilse2018attention}. While these approaches capture broad tissue context, MiTHras focuses on representations within cell-centered patches, combining image-level class information with local token labels. Pretraining methods also differ in how they incorporate supervision. General-purpose encoders learn broadly transferable features, whereas supervised approaches like SupMAE \cite{liang2022supmae} incorporate class labels during pretraining. MiTHras bridges these approaches, providing task-specific supervision at both image and token levels over a large, unannotated corpus.}

\subsection{Masked autoencoders} 
The core idea of MAEs is to learn robust representations by reconstructing images from masked or partially observed patches. Inspired by masked language modeling in BERT \cite{devlin2019bert} and by vision transformers (ViTs) \cite{dosovitskiy2020image}, MIM \cite{he2022masked, wei2023diffusion} has gained attention as a generative SSL approach. MAE reconstructs raw pixels through an asymmetric encoder-decoder \cite{he2022masked}, and DiffMAE \cite{wei2023diffusion} extends this with iterative diffusion processes.

\subsection{Supervised contrastive learning}
Contrastive learning typically models the similarity and dissimilarity of two augmented versions of an image, as shown in SimCLR \cite{chen2020simple} or between images and their corresponding captions, as in CLIP \cite{radford2021learning}. In standard contrastive learning, two views of an image form a positive pair. SupCon \cite{khosla2020supervised} reformulated contrastive learning into multi-positive pair contrastive learning by defining all images with the same label as positive samples. SupMAE \cite{liang2022supmae} further extended the supervised setting into MIM by incorporating label-based supervision and reconstruction-based self-supervision.

\subsection{Semi-supervised learning} 

Semi-supervised learning leverages large amounts of unlabeled data alongside limited annotations. Common strategies include pseudo-labeling \cite{lee2013pseudo} and consistency regularization \cite{tarvainen2017mean}, which encourage stable predictions under perturbations. Recent approaches combine these techniques with confidence-based filtering.
For instance, FixMatch \cite{sohn2020fixmatch} combined pseudo-labeling with confidence thresholds and consistency regularization, using a fixed threshold to retain only high-confidence predictions. FlexMatch \cite{zhang2021flexmatch} extended this by dynamically adjusting the confidence threshold per class, and SimMatch \cite{zheng2022simmatch} added representation-level consistency, encouraging similar embeddings for augmented views of the same image.
\trinh{MiTHras uses a teacher model trained on annotated MFs to assign pseudo labels to an unlabeled pretraining corpus. These pseudo labels supervise contrastive learning at both image and token levels--a process we term semi-supervised contrastive learning (SCL)--alongside self-supervised masked reconstruction. 
}

\section{Method}\label{section:methods}

\begin{figure}[t!]  \begin{center}
\begin{tabular}{c} 
\includegraphics[width=\columnwidth]{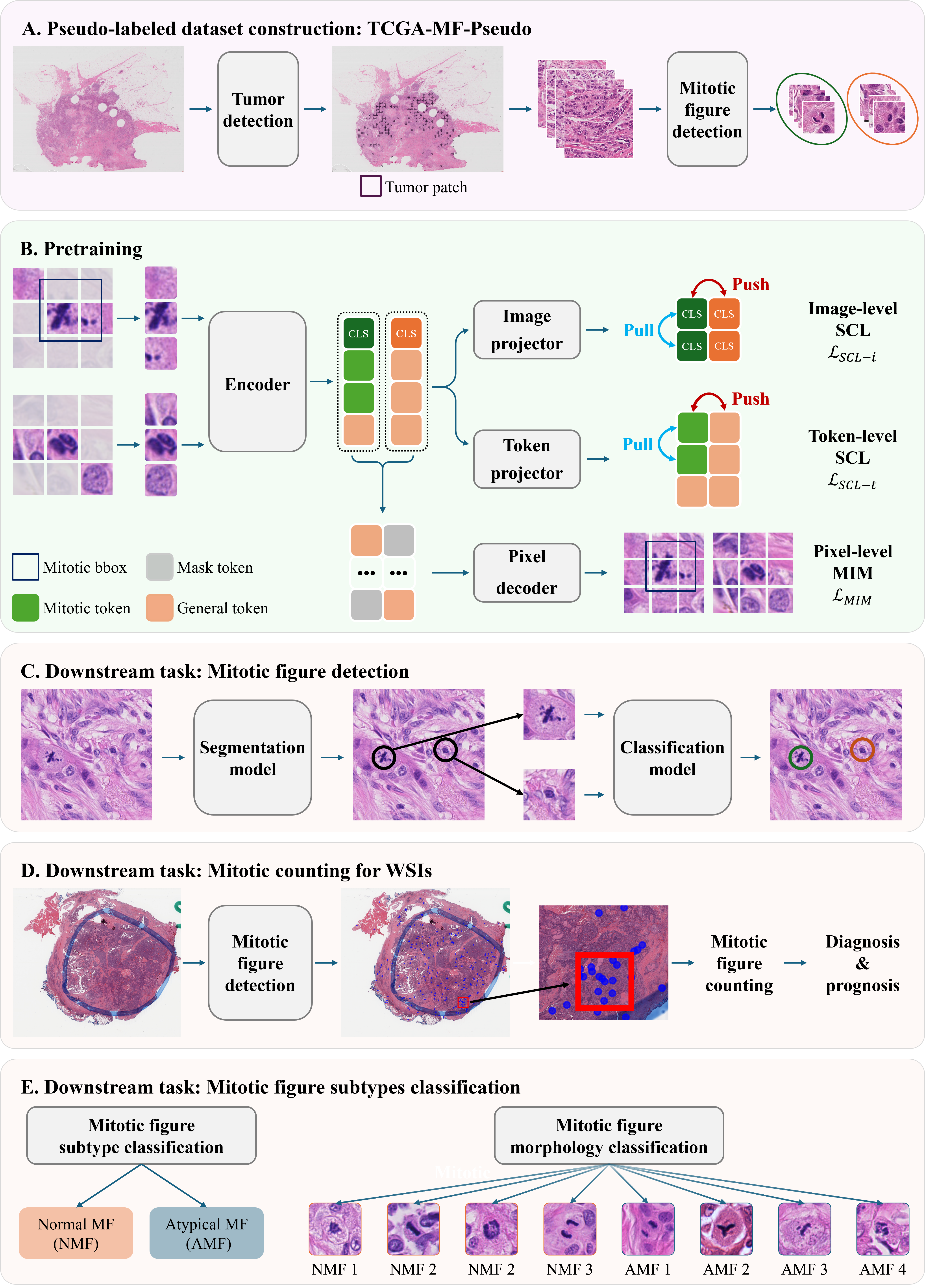}
\end{tabular}
\end{center}
\caption[example]  
{ \label{fig:model} 
Overview of the MiTHras framework.
(A) Pseudo-labeled dataset construction from WSIs, generating mitotic and mimicker samples.
(B) Pretraining with a masked autoencoder and hierarchical contrastive learning at image and token levels.
(C) Two-stage mitotic detection via segmentation and classification.
(D) WSI-level mitotic counting.
(E) Mitotic figure subtype classification.
}
\end{figure}

In this section, we introduce \textbf{MiTHras}, a task-specific hierarchical semi-supervised contrastive masked autoencoder for MF analysis, illustrated in Figure~\ref{fig:model}. We first construct the pseudo-labeled dataset TCGA-MF-Pseudo (Figure~\ref{fig:model}A) and pretrain the encoder \trinh{$f^{\text{enc}}$} on it with a masked autoencoder and task-specific contrastive objectives (Figure~\ref{fig:model}B). On that encoder we build a classification model \trinh{$f^{\text{cls}} = f^{\text{head}} \circ f^{\text{enc}}$} for MFs versus mimickers (Figure~\ref{fig:model}C-right) and for subtypes (Figure~\ref{fig:model}E); a two-stage detection model that pairs a segmentation model $f^{\text{seg}}$ proposing candidates with $f^{\text{cls}}$ (Figure~\ref{fig:model}C); and a whole-slide mitotic count used for survival prediction (Figure~\ref{fig:model}D).
\trinh{Table~\ref{tab:notation} summarizes the notation. Upright superscripts (e.g., $\text{enc}$, $\text{img}$, $\text{tok}$) denote module or level identifiers, italic capitals denote cardinalities, and bold symbols denote vectors or sets of vectors.}

\begin{table}[t]
\centering
\caption{\trinh{Summary of notation.}}
\label{tab:notation}
\begin{adjustbox}{width=0.48\textwidth}
\trinh{\begin{tabular}{ll}
\toprule
Symbol & Description \\
\midrule
$N$, $T$ & Images per batch; tiles per image \\
$V$, $M$ & Visible and masked tiles per image ($V + M = T$) \\
$K$ & Visible tokens per multiviewed batch ($K = 2NV$) \\
$D_E$, $D_P$ & Encoder ($768$) and projection ($512$) dimensions \\
$\alpha$, $\beta$ & Masking ratio; weight of the image-level SCL term \\
\midrule
$f^{\text{enc}}$, $f^{\text{dec}}$ & ViT encoder and MIM decoder \\
$f^{\text{img}}$, $f^{\text{tok}}$ & Image-level and token-level projection heads \\
$f^{\text{head}}$ & Linear classification head \\
$f^{\text{cls}}$, $f^{\text{seg}}$ & Classification ($f^{\text{head}} \circ f^{\text{enc}}$) and segmentation models \\
$f^{\text{det}}$ & Detection model, $f^{\text{det}} = f^{\text{cls}} \circ f^{\text{seg}}$ \\
$f^{\text{cls-br}}$, $f^{\text{cls-morph}}$ & Subtype and morphology classification models \\
\midrule
$\bm{p}$, $\bm{t}$, $\bm{h}$, $\bm{z}$ & Tile, token, encoder output, projected embedding \\
$\mathcal{T}$, $\bm{Z}$ & Set of tokens; set of projected embeddings \\
$\hat{y}$, $\hat{y}^{\text{tok}}$ & Image-level and token-level pseudo labels \\
$P(u)$ & Positive set of anchor $u$: samples sharing its pseudo label \\
$s^{\text{seg}}$, $s^{\text{cls}}$ & Segmentation and classification confidence of a candidate \\
$i$, $u$, $j$, $k$, $w$ & Image, view, tile, token, and batch-token indices \\
$q$ & Index over the positive set in the contrastive sums \\
\bottomrule
\end{tabular}}
\end{adjustbox}
\end{table}

\subsection{Pseudo-label dataset construction} 
\label{Pseudo label dataset construction}

\trinh{General-purpose pathology pretraining commonly uses tissue patches sampled without selecting a particular cell type. MF analysis instead requires discrimination of individual MFs from visually similar non-mitotic objects, or \emph{mimickers}. We construct TCGA-MF-Pseudo to provide cell-centered crops sourced from 14 TCGA cohorts spanning 11 organ sites, with variations in tumor type, scanner, and staining protocols. The construction procedure has three stages:}

\textbf{1) Patch Extraction:} We extract tissue patches of size 896$\times$896 pixels using Otsu's method \cite{otsu1975threshold}. \trinh{Each extracted tile provides tissue context and is subdivided into a $4 \times 4$ grid of $224 \times 224$ patches for tumor filtering. The same tile also accommodates the $512 \times 512$ crops used by the segmentation stage.}

\textbf{2) Tumor Patch Selection:} We select tumor patches if all 4$\times$4 patches of size 224$\times$224 pixels are predicted as tumors using IMPaSh \cite{vuong2022impash}, a model classifying histology textures into adipose, background, debris, lymphocytes, mucus, smooth muscle, normal colon mucosa, cancer-associated stroma, and colorectal adenocarcinoma epithelium (tumor).

\textbf{3) Pseudo-Label Assignment.}
We detect and classify MF and mimicker candidates with the two-stage model of \cite{jahanifar2024mitosis} trained on MIDOG22, replacing its stage-2 classifier with a ViT-base pretrained by MAE on ImageNet. From each refined centroid a 50$\times$50 pixel pseudo bounding box is generated, following the MIDOG22 procedure. \trinh{Candidates are selected via segmentation ($s^{\text{seg}}$) and classification ($s^{\text{cls}}$) scores. Pseudo MFs require $s^{\text{seg}} \geq 0.1$ and $s^{\text{cls}} > 0.5$ (660,831 crops), whereas a pseudo \emph{mimicker} requires $s^{\text{seg}} \geq 0.5$ and $s^{\text{cls}} < 0.5$ (1,133,624 crops). Candidates with $0.1 \leq s^{\text{seg}} < 0.5$ and $s^{\text{cls}} < 0.5$ are excluded. 
The lower segmentation threshold retains potential MFs, relying on the classifier to filter false positives. The higher threshold for mimickers selects candidates with strong segmentation responses that the classifier labels as non-mitotic. Pseudo-label quality is evaluated in Section~\ref{sec:pseudo-validity}.
}

Following this procedure, we obtain a task-specific MF pretraining dataset, referred to as \textbf{TCGA-MF-Pseudo}, which comprises 660,831 pseudo MF images and 1,133,624 pseudo mimicker images. Figure \ref{fig:dataset} illustrates the proportions of the datasets used in this study. Compared to other datasets, TCGA-MF-Pseudo is not only larger in terms of MFs and mimicker images but also covers a broader range of tumor types and data sources.

To assess the effectiveness of task-specific pretraining, we also construct an unlabeled pathology pretraining dataset, referred to as \textbf{TCGA-Unlabeled}, following the standard SSL approach in computational pathology. This dataset consists of 8 million patches (224$\times$224 pixels) extracted from the same \trinh{14 cohorts} at multiple magnifications: 3 million patches at 40$\times$, 2.7 million patches at 20$\times$, 1.2 million patches at 10$\times$, and 1.1 million patches at 5$\times$ magnification.

\trinh{We note that IMPaSh was trained to recognize colorectal tissue categories and is applied here across multiple organ sites. Its role is to select tissue regions, while the separate MIDOG22-trained detector and classifier assign MF and mimicker labels. 
}

\subsection{MiTHras pretraining}
Figure~\ref{fig:model}B details the architecture of the MiTHras pretraining procedure, which comprises five main components: 

\textbf{1) Augmentation and Masking}: Each image is randomly augmented and partially masked to generate two different views composed of visible and masked image tiles.  

\textbf{2) Encoder (\trinh{$f^{\text{enc}}$})}: The visible tiles are passed through a ViT encoder to generate latent feature representations, including a [CLS] token that summarizes the image-level information.

\textbf{3) SCL-global module}: An image-level projection head \trinh{$f^{\text{img}}$} extracts global embeddings from the encoder output, which are trained to align with others from the same pseudo-label class while being distinct from those of different classes.

\textbf{4) SCL-local module}: A separate token-level \trinh{projection head} \trinh{$f^{\text{tok}}$} captures fine-grained, tile-level features. Tokens sharing the same pseudo-label are encouraged to cluster together.

\textbf{5) Decoder (\trinh{$f^{\text{dec}}$})}: A lightweight decoder reconstructs the original image from visible and masked tokens, using a pixel-wise loss to enforce reconstruction quality.

The following subsections describe each component in detail and explain how they are integrated within the MiTHras framework.

\subsubsection{Image masking and encoding} \label{Image_masking_and_encoding}
\trinh{Suppose that we are given a training batch $B = (\bm{X}, \hat{\bm{Y}})$ from TCGA-MF-Pseudo where $\bm{X} = \{\bm{x}_i\}_{i=1}^{N}$ denotes a set of images, $\hat{\bm{Y}} = \{(\hat{y}_i,  \hat{\bm{y}}_i^{\text{bbox}})\}_{i=1}^{N}$ represents their corresponding pseudo labels, $\hat{y}_i$ denotes the pseudo class label for the image $\bm{x}_i$, $\hat{\bm{y}}_i^{\text{bbox}}$ is the set of pseudo bounding boxes for MFs in $\bm{x}_i$, if any are present, and $N$ is the batch size.}

Following the patching process in ViT \cite{dosovitskiy2020image}, an image $\bm{x}_i$ is divided into a set of non-overlapping tiles \trinh{$\{\bm{p}_{i,j}\}_{j=1}^{T}$, where $T$ is the total number of tiles per image.} A large portion of these tiles, determined by a masking ratio $\alpha$, is randomly masked out (i.e., removed from the input), resulting in a subset of visible tiles \trinh{$\{\bm{p}^v_{i,j}\}_{j=1}^{V}$} and a subset of masked tiles \trinh{$\{\bm{p}^m_{i,j}\}_{j=1}^{M}$, where $V$ is the number of visible tiles, $M$ is the number of masked tiles, $V + M = T$}, and $\alpha$ is set to 0.75, following \cite{he2022masked}.
\trinh{Each visible tile $\bm{p}^v_{i,j}$ is flattened and projected into a $D_E$-dimensional embedding space, producing visible tokens $\{\bm{t}^v_{i,k}\}_{k=1}^{V}$. Positional embeddings are added to these visible tokens, and a special [CLS] token $\bm{t}_{i}^{\text{cls}}$ is appended to these tokens, resulting in the input tokens $\{ \bm{t}_{i}^{\text{cls}}, \bm{t}^v_{i,1}, \cdots, \bm{t}^v_{i,V} \}$. These input tokens are fed into the encoder to produce their latent representations $f^{\text{enc}}(\{ \bm{t}^{\text{cls}}_{i}, \bm{t}^v_{i,1}, \cdots, \bm{t}^v_{i,V} \}) = \{\bm{h}_{i,k}^v\}_{k=0}^{V} \in \mathbb{R}^{(V + 1) \times D_E}$, where $D_E = 768$, $\bm{h}^v_{i,0}$ is the embedding for the [CLS] token, and $\{\bm{h}^v_{i,k}\}_{k=1}^{V}$ are the visible token embeddings.}

\subsubsection{Hierarchical semi-supervised contrastive learning branch}
\label{SCL_loss}
\trinh{The hierarchical SCL branch uses pseudo labels to define which representations should be similar. Image-level labels distinguish MF from mimicker crops. Token level labels derived from pseudo bounding boxes separate putative MF regions from surrounding tissue. These two objectives encourage class separation at different spatial scales without requiring manual annotation of the pretraining corpus.}

Each image $\bm{x}_i$ is randomly augmented twice to form a multiviewed batch that contains $2N$ augmented images indexed by $u \in U \equiv \{1,\cdots,2N \}$. \trinh{Each view $\bm{x}_u$ is masked out and encoded by $f^{\text{enc}}$ (Section~\ref{Image_masking_and_encoding}), and the resulting hidden states $\{\bm{h}_{u,k}^v\}_{k=0}^{V}$ are mapped by a projection head into a $D_P$-dimensional space ($D_P = 512$), where index $k=0$ denotes the [CLS] position and $k \geq 1$ the visible tokens. The head is instantiated separately at the two levels, as $f^{\text{img}}$ and $f^{\text{tok}}$ below.}
\trinh{Given a collection of $\ell_2$-normalized embedding vectors $\{\bm{z}_u\}_{u \in U}$, one per element to be contrasted, a} semi-supervised contrastive loss \trinh{$\mathcal{L}_{\text{SCL}}$} can be defined as follows:
\begin{equation}
  \trinh{\mathcal{L}_{\text{SCL}}(\bm{X},\hat{\bm{Y}}) =  \frac{0.1}{2N}\sum\limits_{u \in U} \frac{-1}{|P(u)|} \sum\limits_{q \in P(u)}\log \frac{\exp(\bm{z}_u \cdot  \bm{z}_{q}/\tau)}{\sum\limits_{a \in A(u)} \exp(\bm{z}_u\cdot \bm{z}_a/\tau)}}
\end{equation}
where $A(u) \equiv  U \setminus \{u\}$, \trinh{$P(u) \equiv \{q \in A(u): \hat{y}_q=\hat{y}_u\}$  is the set of positive indices sharing the same pseudo label as $\bm{x}_u$}, $|P(u)|$ is the cardinality of $P(u)$, and $\tau=0.1$ is the temperature. \trinh{The loss averages over anchors and includes a fixed multiplier of $0.1$. Both the image- and token-level objectives use this temperature, scaling, and embedding normalization.}
Building upon $\mathcal{L}_{\text{SCL}}$, we define a hierarchical contrastive loss that combines image-level (\trinh{$\mathcal{L}_{\text{SCL}}^{\text{img}}$}) and token-level (\trinh{$\mathcal{L}_{\text{SCL}}^{\text{tok}}$}) objectives.

\subsubsection{Image-level semi-supervised contrastive learning}

To improve the quality of latent representations at the global (image) level, we introduce image-level SCL. Specifically, given an image view $\bm{x}_u$, we adopt an image-level projection head \trinh{$f^{\text{img}}$} to produce the embedding vectors \trinh{$\bm{Z}^{\text{img}}_{u} = \{\bm{z}^{\text{img}}_{u,k}\}_{k=0}^{V} = \{ f^{\text{img}} (\bm{h}_{u,k}^v) \}_{k=0}^{V} \in \mathbb{R}^{(V + 1) \times D_P}$} as described in Section~\ref{SCL_loss}.
Using the embedding vectors, we define the image-level contrastive loss (\trinh{$\mathcal{L}_{\text{SCL}}^{\text{img}}$}) given by:

\begin{equation}
  \trinh{  \mathcal{L}_{\text{SCL}}^{\text{img}}(\bm{X},\hat{\bm{Y}}) =
  \frac{0.1}{2N}\sum_{u \in U} \frac{-1}{|P(u)|}
  \sum_{q \in P(u)}\log
  \frac{\exp(\bm{z}^{\text{cls}}_u \cdot  \bm{z}^{\text{cls}}_{q}/\tau)}
  {\sum\limits_{a\in A(u)} \exp(\bm{z}^{\text{cls}}_u\cdot \bm{z}^{\text{cls}}_a/\tau)}}
\end{equation}
where \trinh{$\bm{z}_u^{\text{cls}} = \bm{z}^{\text{img}}_{u,0}$} is the projected embedding vector for the [CLS] token.

\subsubsection{Token-level semi-supervised contrastive learning}
\label{Token_level_SCL}
At the local (token) level, we promote fine-grained feature learning by applying an SCL loss to individual visible tokens. As described in Section~\ref{Image_masking_and_encoding}, each image view $\bm{x}_u$ in batch $B$ is tokenized into a set of visible tokens \trinh{$\{\bm{t}^v_{u,k}\}^{V}_{k=1}$}. Similar to the image-level SCL, we employ a token-level projection head \trinh{$f^{\text{tok}}$} to obtain projected embeddings \trinh{$\{\bm{z}^{\text{tok}}_{u,k}\}_{k=0}^{V} = \{ f^{\text{tok}} (\bm{h}_{u,k}^v ) \}_{k=0}^{V} \in \mathbb{R}^{(V + 1) \times D_P}$}. We then discard the embedding corresponding to the [CLS] token and retain only the projected visible token embeddings, denoted as \trinh{$\bm{Z}^{\text{tok}}_{u} = \{\bm{z}^{\text{tok}}_{u,k}\}_{k=1}^{V}$}.
For MF images, tokens are labeled from the pseudo bounding boxes \trinh{$\hat{\bm{y}}^{\text{bbox}}_u$}; for mimicker images all tokens are \textit{non-mitotic}. \trinh{Pretraining inputs are extracted as $128 \times 128$ crops, with the central $50 \times 50$ region forming the bounding box. Augmentation proceeds in two stages. The spatial stage begins with a randomly resized crop to $224 \times 224$ pixels with the cropped area fraction uniformly sampled from $[0.2,1.0]$, followed by random horizontal and vertical flips. These spatial transformations are applied jointly to both the image and the bounding box. Then, the photometric stage applies color jitter, random grayscale, Gaussian blur, and solarization for the second view to the image alone. A token is labeled \emph{mitotic} if its corresponding patch cell lies inside the box, with box edges rounded to the nearest patch boundary. No separate boundary class is used.}

\trinh{For a square crop, the sampled area fraction corresponds to a linear magnification of approximately $1.75$--$3.9\times$. Before boundary rounding, a fully retained $50\times50$ box occupies $30$--$149$ patch-cell areas in the $14\times14$ token grid. Actual token labels depend on the crop position, aspect ratio, and boundary rounding; partial cropping reduces box coverage, and a crop that excludes the box contains no mitotic tokens. The fixed box may include adjacent tissue or exclude larger figures, yielding only coarse spatial supervision.}
This results in a set of visible tokens with corresponding pseudo labels for the entire training batch $B$, denoted as \trinh{$B^{\text{tok}} = (\mathcal{T}, \hat{\bm{Y}}^{\text{tok}})$ where $\mathcal{T} = \{ \bm{t}^v_w \}_{w=1}^{K}$ denotes all visible tokens in $B$, $\hat{\bm{Y}}^{\text{tok}} = \{ \hat{y}^{\text{tok}}_w \}_{w=1}^{K}$ denotes their corresponding pseudo labels, $\hat{y}^{\text{tok}}_w$ denotes a pseudo label for the corresponding token $\bm{t}^v_w$, and $K = 2 \times N \times V$ is the total number of visible tokens in $B$.}
Using these tokens and their corresponding pseudo labels, we compute the token-level SCL (\trinh{$\mathcal{L}_{\text{SCL}}^{\text{tok}}$}), defined as:

\begin{equation}
  \begin{split}
  \trinh{\mathcal{L}_{\text{SCL}}^{\text{tok}}(\mathcal{T},\hat{\bm{Y}}^{\text{tok}}) =
  \frac{0.1}{K}\sum_{w \in W} \frac{-1}{|P^{\text{tok}}(w)|}
  \sum_{q \in P^{\text{tok}}(w)}} \\
  \trinh{\log \frac{\exp(\bm{z}^{\text{tok}}_w \cdot  \bm{z}^{\text{tok}}_{q}/\tau)}
  {\sum\limits_{a\in A^{\text{tok}}(w)} \exp(\bm{z}^{\text{tok}}_w \cdot \bm{z}^{\text{tok}}_a/\tau)}}
  \end{split}
\end{equation}
\trinh{where $W \equiv \{1,\cdots,K \}$, $A^{\text{tok}}(w) \equiv W \setminus \{w\}$, $P^{\text{tok}}(w) \equiv \{q \in A^{\text{tok}}(w): \hat{y}^{\text{tok}}_q = \hat{y}^{\text{tok}}_w\}$ is the set of positive indices sharing the same pseudo label as $\bm{t}^v_w$, and $|P^{\text{tok}}(w)|$ is the cardinality of $P^{\text{tok}}(w)$.}

\subsubsection{Reconstruction branch}
We add a pixel-level MIM branch. For an image view $\bm{x}_u$, masked tile positions are replaced by learnable mask tokens \trinh{$\{ \bm{t}^m_{u,j} \}_{j=1}^{M}$} \cite{devlin2019bert}, decoder-specific positional embeddings are added to the visible and mask tokens, and the combined set is passed to a lightweight MIM decoder \trinh{$f^{\text{dec}}$} that predicts the pixel content of the masked tiles. The reconstruction $\bm{x}^r_u$ comprises the visible tiles \trinh{$\bm{p}^{v}_{u} = \{ \bm{p}^{v}_{u,j} \}_{j=1}^{V}$} and the recovered masked tiles \trinh{$\bm{p}^{\text{rec}}_{u} = \{ \bm{p}^{\text{rec}}_{u,j} \}_{j=1}^{M}$}.
\trinh{To optimize $f^{\text{dec}}$, we minimize the mean squared error (MSE) between reconstructed masked tiles $\bm{p}^{\text{rec}}_{u,j}$ and their normalized pixel targets $\tilde{\bm{p}}^{m}_{u,j}$. The MSE averages over the pixel and channel values within each tile, and the reconstruction loss averages over the set $\mathcal{M}$ of masked tiles:}

\begin{equation}
  \trinh{\mathcal{L}_{\text{MIM}}(\bm{X}^r, \bm{X}) = \underset{(u,j)\in\mathcal{M}}{\operatorname{mean}}\operatorname{MSE}\!\left(\bm{p}^{\text{rec}}_{u,j},\tilde{\bm{p}}^{m}_{u,j}\right).}
\end{equation}

\subsubsection{MiTHras loss function}
Overall, the proposed MiTHras framework is optimized using the following objective function:
\begin{equation}
\begin{aligned}
\trinh{\mathcal{L}(\bm{X}, \bm{X}^r, \mathcal{T}, \hat{\bm{Y}}, \hat{\bm{Y}}^{\text{tok}})} &= \trinh{\mathcal{L}_{\text{MIM}}(\bm{X}^r, \bm{X})  + \beta \mathcal{L}_{\text{SCL}}^{\text{img}}(\bm{X}, \hat{\bm{Y}})}  \\
&\quad\trinh{+ (1-\beta) \mathcal{L}_{\text{SCL}}^{\text{tok}}(\mathcal{T}, \hat{\bm{Y}}^{\text{tok}}).}
\end{aligned}
\end{equation}
\trinh{The parameter $\beta$ weights image-level SCL, while $1-\beta$ weights token-level SCL and the reconstruction term has unit weight. The main MiTHras configuration uses $\beta=0.75$. The ablation at $\beta=1$ combines masked reconstruction with image-level SCL and excludes token-level SCL; $\beta=0$ combines reconstruction with token-level SCL and excludes image-level SCL. }

\begin{figure*} [!t]
  \begin{center}
  \begin{tabular}{c} 
\includegraphics[width=0.95\textwidth]{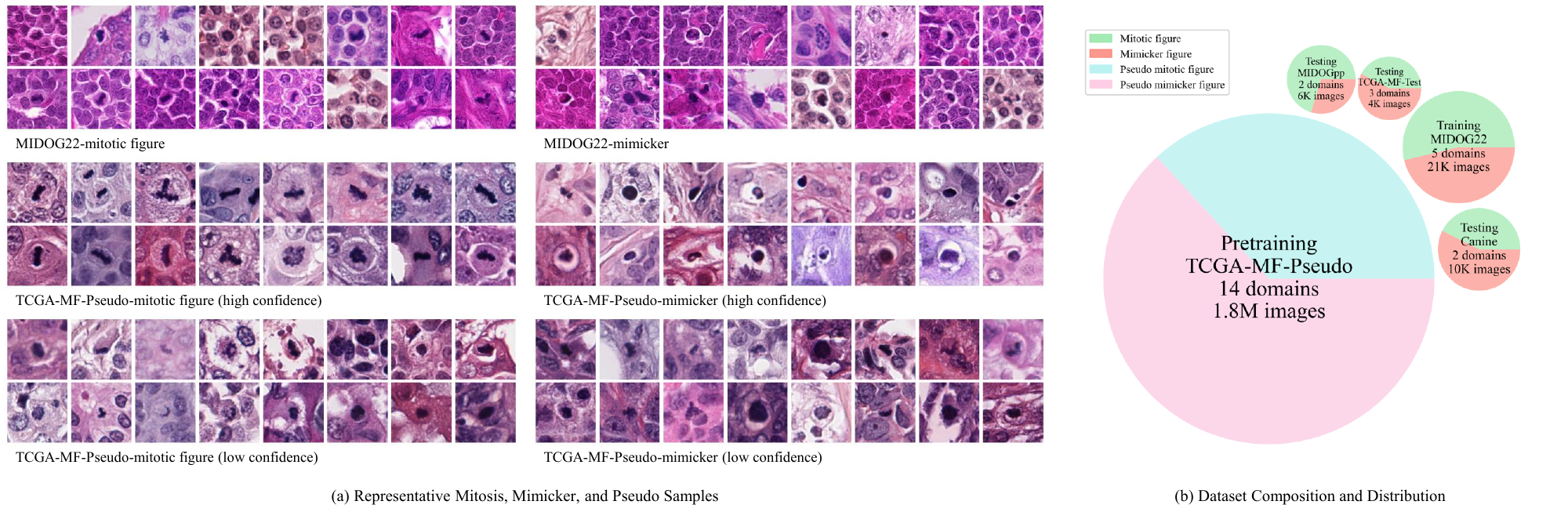}
  \end{tabular}
  \end{center}
  \caption[example]  
  { \label{fig:dataset} 
(a) Representative mitosis, mimicker, and pseudo samples.
(b) Dataset distribution shown as a bubble chart, where bubble size reflects the number of images and labels indicate domain count and dataset size. TCGA-MF-Pseudo is the largest dataset, spanning 14 domains ($\sim$1.8M images).}
  \end{figure*} 

\subsubsection{Architecture}
The encoder \trinh{$f^{\text{enc}}$} is a standard ViT-base \cite{dosovitskiy2020image}: 12 transformer blocks, embedding dimension \trinh{$D_E = 768$}, 12 attention heads. The MIM decoder \trinh{$f^{\text{dec}}$} has 8 blocks with dimension 512. The image and token projection heads \trinh{$f^{\text{img}}$} and \trinh{$f^{\text{tok}}$} share one configuration, a fully connected layer mapping \trinh{$D_E$}- to \trinh{$D_P$}-dimensional embeddings for both the class token and the tile tokens. Downstream, only \trinh{$f^{\text{enc}}$} is retained; \trinh{$f^{\text{dec}}$}, \trinh{$f^{\text{img}}$} and \trinh{$f^{\text{tok}}$} are discarded.

\subsection{Mitotic figure classification model} \label{classification_model}
The classification model $f^{\text{cls}}$ serves both the mitotic-versus-mimicker task and subtype classification. We append a classification head \trinh{$f^{\text{head}}$} to the encoder \trinh{$f^{\text{enc}}$}, giving \trinh{$\bm{y}'=f^{\text{cls}}(\bm{x})= f^{\text{head}}(f^{\text{enc}}(\bm{x}))$}. The encoder is initialized from the MiTHras pretrained weights and the head with zero weights and a constant bias; all parameters are fine-tuned end-to-end with cross-entropy loss.

\subsection{Mitotic figure detection model}

\trinh{The detection model follows the two-stage pipeline of Jahanifar \textit{et al.} \cite{jahanifar2024mitosis}. A segmentation model proposes candidate locations, and the MF-versus-mimicker classifier evaluates a crop around each candidate. We write this pipeline as $f^{\text{det}}=f^{\text{cls}}\circ f^{\text{seg}}$, with candidate extraction and cropping included between the two stages.}

In the first stage, we use an \trinh{Efficient-UNet-b0} segmentation model $f^{\text{seg}}$ to process an input image and generate a probability map, highlighting candidate regions for mitotic figures. Candidate MFs are then identified by applying morphological post-processing to this map. For each candidate, the centroid is extracted, and the mean intensity within the region is used as the detection \trinh{score $s^{\text{seg}}$}. In the second stage, each candidate is classified using the mitotic classification model $f^{\text{cls}}$, as described in Section~\ref{classification_model}, yielding a classification score \trinh{$s^{\text{cls}}$}. \trinh{The stages are trained independently. For each two-stage method, a single pair of segmentation and classification thresholds is selected by maximizing F$_1$ across the three MIDOG22 validation folds and is then held fixed for external evaluation. The candidate detector is shared across second-stage classifiers.}

\trinh{During inference, the segmentation model processes $512\times512$ windows with a stride of 400 pixels, and overlapping probability maps are averaged. The 112-pixel overlap reduces boundary effects while limiting repeated computation. This setting follows the MDFS protocol \cite{jahanifar2024mitosis}; its approximate processing redundancy is $(512/400)^2=1.64$, compared with $4$ for a half-window stride.}

\subsection{MiTHras mitotic count for WSIs}
\trinh{We apply the two-stage detector to WSIs and summarize the detections using a hotspot count in $2\,\mathrm{mm}^2$ and a density expressed per $20\,\mathrm{mm}^2$. The hotspot summarizes the most active sampled region, whereas density summarizes average activity across the analyzed tissue. These count-derived features are evaluated for survival prediction on TCGA-THCA and TCGA-GBMLGG.}
For both the mitotic hotspot (2 mm$^2$) and mitotic density (20 mm$^2$) strategies, we begin by applying Otsu \cite{otsu1975threshold} thresholding to extract tissue patches from a WSI. The patches are of size 512$\times$512 pixels acquired at 40$\times$ magnification (0.25 $\mu$m per pixel). Each patch is then processed using the two-stage mitotic detection pipeline to identify MFs. 

\subsubsection{Mitotic hotspot (2 mm$^2$)} 

\trinh{Local Max ROI Counting (LMRC) centers a square of area $2\,\mathrm{mm}^2$ on each detected MF and returns the largest count among these candidate regions. Detected coordinates are indexed with a KD-tree (\texttt{cKDTree}, SciPy v1.11.2). A Chebyshev-distance query retrieves points within each square using a half-side length of $\sqrt{2}/2$ mm. The procedure avoids evaluating a dense sliding grid; its search is restricted to detection-centered squares.}

\subsubsection{Mitotic density (20 mm$^2$)}

\trinh{Mitotic density expresses the total detected MF count per unit of analyzed tissue area, with $20\,\mathrm{mm}^2$ used as the reporting reference. Unlike the hotspot count, it summarizes average activity across the sampled tissue.}

\subsection{Mitotic figure subtype classification model} \label{classification_subtypes_model}
Similar to the MF classification model $f^{\text{cls}}$, we build two subtype classification models: $f^{\text{cls-br}}$ for the binary classification of mitotic figure subtypes and $f^{\text{cls-morph}}$ for the 8-class morphological subtype classification.

\section{Experiments} \label{section:experiment}

\subsection{Datasets}
\label{sec:datasets}
We evaluate four tasks, mitotic figure classification, detection, survival prediction from mitotic count, and subtype classification, across the datasets detailed in Table~\ref{tab:models-datasets} and Figure~\ref{fig:dataset}.
\trinh{The TCGA evaluation cancer types (GBM, SARC, DLBC, THCA, and LGG) are excluded from TCGA-MF-Pseudo. TCGA pretraining and evaluation therefore share no slide, patient, or cancer type. TCGA metadata show no overlap in tissue source site codes, although 20 of the 33 parent institutions contributing to the evaluation cohorts also contribute to the pretraining cohorts. The TCGA split therefore evaluates transfer across cancer types but does not establish independence of institutional acquisition practices. MIDOGpp, Canine, AMi-Br-TUPAC, and AMi-Morph-TUPAC provide additional external evaluations across laboratories, scanners, and, for Canine, species.}

\subsubsection{Pretraining and training data}
TCGA-MF-Pseudo (Section~\ref{Pseudo label dataset construction}) provides 660,831 pseudo MFs and 1,133,624 pseudo mimickers for pretraining. For supervised training we adopt \textbf{MIDOG22} \cite{aubreville2024domain}, containing \trinh{9,501 MFs and 11,051 mimickers from 354 images across five tumor types: 150 human breast carcinomas, 44 canine lung carcinomas, 55 canine lymphosarcomas, 50 canine cutaneous mast cell tumors, and 55 human neuroendocrine tumors.} \trinh{Following the three-fold MDFS training protocol \cite{jahanifar2024mitosis}, we train Efficient-UNet-b0 for candidate segmentation and reuse this segmentation stage for all second-stage classifier comparisons. The three train/validation splits contain 239/115, 234/120, and 235/119 source images, respectively, with no image overlap between training and validation within a split. We extract $512\times512$ pixel patches for segmentation and $128\times128$ pixel crops centered on MFs and mimickers for classification, and select the best classification checkpoint per fold by validation performance.} \trinh{Each fold is trained and evaluated independently. Results are reported as the mean $\pm$ standard deviation across three folds, without ensembling predictions. All methods use the same folds. The standard deviation describes variation across training folds.}

\subsubsection{Tasks 1 and 2: classification and detection}
Both tasks are evaluated on MIDOGpp, Canine, and TCGA-MF-Test. \textbf{MIDOGpp} contains 2,436 MFs and 3,300 mimickers from \trinh{100 canine soft tissue sarcoma and 49 human melanoma images.} \textbf{Canine} includes 7,206 MFs and 3,082 mimickers from 160 canine cutaneous mast cell tumor and 105 canine mammary tumor images \cite{aubreville2020completely, bertram2019large}. \textbf{TCGA-MF-Test} comprises 1,539 MFs and 2,073 mimickers from TCGA glioblastoma multiforme (GBM), soft tissue sarcoma (SARC), and lymphoid neoplasm diffuse large B-cell lymphoma (DLBC).

For classification, we extract $128 \times 128$ pixels centered on annotated centroids. \trinh{Evaluation metrics are defined in Section~\ref{sec:evaluation-metrics}.} For detection, we use full images: segmentation probability maps are produced via a $512 \times 512$ sliding window at stride 400 and aggregated by averaging overlaps. Peak points define candidate locations, from which $128 \times 128$ patches are extracted and classified. Predictions are matched to ground truth within a 30-pixel radius and scored by F$_1$.

\subsubsection{Task 3: Survival prediction using mitotic count}
\label{sec:task3}
\trinh{Given the role of mitotic count in grading high-grade follicular-cell-derived thyroid carcinoma \cite{jung2022update} and diffuse glioma, we evaluate its association with overall survival across two cohorts: TCGA-THCA (406 thyroid WSIs from 395 patients, including 11 deaths) and TCGA-GBMLGG (844 brain WSIs from 501 patients, including 122 deaths). The slides were scanned at $40\times$ magnification. Survival time is measured in months, with death coded as the event ($\mathrm{event}=1-\mathrm{censorship}$).} 

\subsubsection{Task 4: Mitotic figure subtype classification}
We evaluate generalization on two subtype benchmarks \cite{bertram2025histologic}: \textbf{AMi-Br} (typical vs.\ atypical) and \textbf{AMi-Morph} (eight morphology classes), used to evaluate $f^{\text{cls-br}}$ and $f^{\text{cls-morph}}$ respectively. Three-fold cross-validation is performed on AMi-Br-MIDOG21 and AMi-Morph-MIDOG21. \trinh{Each fold model is applied independently to the held-out AMi-Br-TUPAC and AMi-Morph-TUPAC domains, and results are reported as the mean $\pm$ standard deviation over the three folds.} AMi-Br-MIDOG21 contains 1,721 figures (1,317 typical, 404 atypical); AMi-Morph-MIDOG21 adds eight-way annotations to the same images from two experts, of which we use expert~1 (four typical subtypes: prometaphase, metaphase, ring-shaped, anaphase-telophase; four atypical: bipolar asymmetric, multipolar, segregation-related, other), with the rarest class holding 18 examples. The test sets are AMi-Br-TUPAC (1,999 figures; 1,571 typical, 428 atypical) and AMi-Morph-TUPAC (1,999 figures across the same eight subtypes). AMi-Br is scored by positive-class F$_1$ for atypical figures, and AMi-Morph by macro F$_1$ over the morphology classes.

\begin{table}[t]
\centering
\caption{\trinh{Training and test datasets for the MiTHras models. The classification encoders are pretrained on TCGA-MF-Pseudo and then fine-tuned on the listed training sets. Detection and mitotic counting share the two-stage pipeline $f^{\text{det}} = f^{\text{cls}} \circ f^{\text{seg}}$, trained on MIDOG22.}}
\label{tab:models-datasets}
\begin{adjustbox}{max width=\columnwidth}
\begin{tabular}{ll}
\toprule
\textbf{Task} & \textbf{Train / Test} \\
\midrule
MF vs.\ mimicker classification  & MIDOG22 / MIDOGpp, Canine, TCGA-MF-Test \\
MF detection                     & MIDOG22 / MIDOGpp, Canine, TCGA-MF-Test \\
Mitotic counting on WSIs         & MIDOG22 / TCGA-THCA, TCGA-GBMLGG \\
MF subtype (typical/atypical)    & AMi-Br-MIDOG21 / AMi-Br-TUPAC \\
MF morphology (8 classes)        & AMi-Morph-MIDOG21 / AMi-Morph-TUPAC \\
\bottomrule
\end{tabular}
\end{adjustbox}
\end{table}

\subsection{Implementation details} \label{section:implementation}

We trained all the pretraining models for 200 epochs on 8 NVIDIA A6000 GPUs. \trinh{MiTHras was initialized from scratch and optimized with AdamW (betas $0.9$ and $0.95$, weight decay $0.05$), an effective batch size of $512$, and a learning rate of $3\times10^{-4}$, with 10 warm-up epochs followed by cosine decay. The configuration without MIM used the same image/token weights ($0.75/0.25$) and optimizer schedule, with an effective batch size of $768$ and learning rate of $4.5\times10^{-4}$. The masking ratio was fixed at $\alpha=0.75$, following MAE \cite{he2022masked}, while the ablations varied $\beta$ to assess the balance of the image- and token-level SCL objectives. During pretraining, two augmented views were generated per image, following the procedure described in Section~\ref{Token_level_SCL}. Views were normalized with ImageNet statistics.}
All classification models were fine-tuned for 50 epochs on a single NVIDIA A6000, with RandAugment plus random erasing, $224 \times 224$ inputs, ImageNet normalization, label smoothing ($\epsilon = 0.1$), AdamW (weight decay 0.05, layer-wise decay 0.75), base learning rate $2.5\times10^{-4}$ with five warm-up epochs, and automatic mixed precision. The segmentation model was trained separately for 50 epochs on 2 A6000 GPUs at $512 \times 512$ with standard augmentation and optimization.
\trinh{
For linear probing, the frozen pretrained encoder was attached to a classification head consisting of an affine-free batch normalization layer followed by a trainable linear layer. The head was trained using LARS with a base learning rate of 0.1, zero weight decay, 10 warm-up epochs, cosine decay, and cross-entropy loss. Images were resized to 224$\times$224 pixels with weak augmentation and ImageNet normalization. Affine-free batch normalization standardized feature scales across backbones. All methods used the same three folds and 50-epoch training budget for linear probing and fine-tuning.}

\subsection{\trinh{Evaluation metrics and statistical analysis}}
{\revisioncolor
\label{sec:evaluation-metrics}
We evaluate performance using accuracy (ACC), precision, recall, F$_1$ score, and area under the receiver operating characteristic curve (ROC-AUC). For mitotic-versus-mimicker classification, MFs are the positive class. For AMi-Br, atypical figures (label 1) are positive and typical figures (label 0) are negative. Binary precision, recall, and F$_1$ are calculated for the positive class. For AMi-Morph, these metrics are calculated separately for each class occurring in the targets or predictions and averaged with equal class weights (unweighted macro averaging); undefined class scores are set to zero. ROC-AUC is not evaluated for AMi-Morph. Detection metrics use predicted locations matched to annotated MFs within a 30-pixel radius.

Survival prediction is evaluated using Harrell's concordance index (C-index), which measures agreement between predicted risk and observed event ordering among comparable pairs, accounting for censoring. Cox proportional-hazards models are fitted and evaluated within each full cohort. Patients with multiple slides contribute multiple observations. 95\% confidence intervals are calculated using 1,000 WSI resamples with fitted risk predictions held fixed, i.e., without refitting the Cox models. Log-rank tests compare groups defined by a cutpoint fixed at the median of the designated training subset. A likelihood-ratio test compares a model containing age and sex with one additionally containing the mitotic-count feature. Reported p-values are unadjusted.

Classification, detection, and subtype results are reported as the mean and sample standard deviation (SD) across three training folds. The blinded-review protocol (Section~\ref{sec:audit-protocol}) specifies precision and uncertainty estimation for the pseudo-label audit.
The main tables report F$_1$ for classification, detection, and subtype classification, and the C-index for survival prediction. Additional classification metrics are provided in Supplementary Table S1. Detection and subtype results are in Supplementary Table S2.
}

\subsection{Comparative experiments and ablation study}
\label{section:comparative_experiments}

\subsubsection{Classification task}
\trinh{Table~\ref{table:results_mitosis_cls_f1} compares MiTHras, a ViT-base pretrained on the IMPaSh-selected TCGA-MF-Pseudo corpus, with the MDFS classifier \cite{jahanifar2024mitosis}, general-purpose encoders \cite{oquab2024dinov, radford2021learning}, pathology-pretrained encoders \cite{wang2022transformer, wang2024pathology, kang2023benchmarking, huang2023visual, ma2025generalizable, chen2024towards, vorontsov2024foundation}, and alternative pretraining objectives \cite{liang2022supmae, huang2023contrastive}. ViT-ImageNet and ViT-TCGA-MF-Pseudo use supervised classification pretraining on their named corpora. MAE-ImageNet, MAE-TCGA-Unlabeled, and MAE-TCGA-MF-Pseudo use masked reconstruction on their named corpora. CMAE uses contrastive masked-autoencoder pretraining on TCGA-MF-Pseudo, while SupMAE uses labeled MIDOG22 images. All encoders are evaluated under full fine-tuning and frozen-encoder linear probing (described in Section~\ref{section:implementation}). 
}

\subsubsection{Detection task}
\trinh{Two-stage detection models follow the MDFS pipeline \cite{jahanifar2024mitosis} and are compared with standalone Efficient-UNet-b0 and YOLOv10 \cite{wang2024yolov10} baselines. Efficient-UNet-b0 is trained under the three-fold MDFS protocol and shared across the two-stage classifier comparisons. The compared methods differ in their second-stage classifier. Each classifier is the fine-tuned model from the classification experiment and receives $224 \times 224$ inputs.}

\subsubsection{WSI survival prediction}
\trinh{The two-stage approaches share tissue extraction, the candidate detector, and count computation, and differ in their second-stage classifiers. A segmentation-only baseline computes counts without second-stage classification.}

\subsubsection{Mitotic figure subtype classification task}
\trinh{Subtype experiments compare the pretrained encoders listed in Table~\ref{table:results_mitosis_subtype} on the AMi benchmarks introduced by Bertram \textit{et al.} \cite{bertram2025histologic}. Each encoder receives a task-specific classification head and is fine-tuned using the shared protocol in Section~\ref{section:implementation}. The ablation uses the same protocol.}

\subsubsection{Ablation experiments}
To assess the contribution of each component, we evaluated masked reconstruction and image-level and token-level SCL individually and in combination (Table~\ref{table:results_mitosis_cls_Ablation}). We compare $\beta=0.75$ and $\beta=1$ under full fine-tuning, linear probing, and subtype classification for three settings: ViT-base and ViT-tiny pretrained on TCGA-MF-Pseudo, and ViT-base pretrained on labeled MIDOG22. The same evaluation protocols are used within each paired
comparison (Table~\ref{table:results_mitosis_cls_Ablation}).

\subsubsection{\trinh{Blinded pathological review}}
\label{sec:audit-protocol}
A single pathologist (P.T.H.H.) conducted a blinded review to assess tissue selection and pseudo-label quality across all 14 cohorts. Only the images were provided, and the pipeline labels and confidence scores were concealed. For tissue selection, IMPaSh and CONCH \cite{lu2024visual} were applied to 10,000 sampled patches per cohort. The reviewer examined 25 patches per cohort from each of three strata: retained by both filters, retained only by IMPaSh, and retained only by CONCH, yielding 1,050 reviewed patches. For pseudo-label quality, the audit included 700 crop reviews, comprising 420 pseudo-mitotic and 280 pseudo-mimicker reviews (30 and 20 per cohort, respectively) from the final pretraining corpus. Each crop was classified as confirmed, not confirmed, or unsure. Precision ranges count unsure verdicts as incorrect at the lower endpoint and correct at the upper endpoint. Per-cohort proportions are reported with Wilson confidence intervals. Tumor-filter stratum estimates pool the reviewed patches; pooled pseudo-mitotic precision is also reported with weights proportional to each cohort's pseudo-mitotic pool size. The additional-figure assessment included 199 pseudo-mimicker crops whose central candidate was classified as non-mitotic. For each crop, the reviewer recorded whether an MF was present elsewhere in the crop. For bounding-box assessment, MFs confirmed during the crop review were assigned to the smallest enclosing square category: 25, 50, 75, or larger than 75 pixels. No additional crops were
sampled for this assessment.


\section{Results} \label{section:Results}
\subsection{Mitotic figure classification results}
\label{sec:classification-results}

\begin{table*}[h]
\begin{center}
\caption{\trinh{Mitotic figure classification results (F$_1$, mean $\pm$ standard deviation over three folds) under full fine-tuning and linear probing. Each fold is trained and evaluated independently, without ensembling predictions. $^{\ddagger}$SupMAE is pretrained on labeled MIDOG22 images. Best value per column in bold.}}
\label{table:results_mitosis_cls_f1}
\setlength{\tabcolsep}{2pt} 
\renewcommand{\arraystretch}{1} 
\begin{adjustbox}{width=\textwidth}
\begin{tabular}{ll|c|c|c|c|c|c|c}
\toprule
\multirow{2}{*}{Pretraining Method} & & \multirow{2}{*}{Network} & \multicolumn{3}{c|}{Fine-tuning} & \multicolumn{3}{c}{Linear Probing} \\
\noalign{\smallskip}
\cline{3-9}
\noalign{\smallskip}
& & &\multicolumn{1}{c|}{MIDOGpp} & \multicolumn{1}{c|}{Canine} & \multicolumn{1}{c|}{TCGA-MF-Test} & \multicolumn{1}{c|}{MIDOGpp} & \multicolumn{1}{c|}{Canine} & \multicolumn{1}{c}{TCGA-MF-Test} \\
\midrule
\multirow{4}{3cm}{\textit{\trinh{Natural-image pretraining}}} & MDFS & EfficientNet-b7 & 0.831$\pm$0.003 & 0.815$\pm$0.010 & 0.757$\pm$0.010 & 0.356$\pm$0.309 & 0.708$\pm$0.103 & 0.570$\pm$0.027 \\

& ViT-ImageNet (Supervised) & ViT-base & 0.824$\pm$0.011 & 0.832$\pm$0.006 & 0.766$\pm$0.004 & 0.354$\pm$0.306 & 0.523$\pm$0.321 & 0.155$\pm$0.059 \\

& MAE-ImageNet & ViT-base & 0.840$\pm$0.006 & 0.840$\pm$0.009 & 0.775$\pm$0.007 & 0.318$\pm$0.075 & 0.614$\pm$0.059 & 0.376$\pm$0.033 \\

& DINOv2 & ViT-base & 0.840$\pm$0.006 & 0.842$\pm$0.007 & 0.778$\pm$0.012 & 0.026$\pm$0.024 & 0.545$\pm$0.472 & 0.442$\pm$0.137 \\

& CLIP & ViT-base & 0.829$\pm$0.006 & 0.834$\pm$0.003 & 0.773$\pm$0.005 & 0.235$\pm$0.311 & 0.440$\pm$0.412 & 0.512$\pm$0.025 \\

\midrule
\multirow{9}{3cm}{\textit{Histopathology pretraining models}} & CTransPath-MoCov3 & Swin-tiny & 0.808$\pm$0.013 & 0.821$\pm$0.011 & 0.736$\pm$0.007 & 0.199$\pm$0.343 & 0.603$\pm$0.063 & 0.548$\pm$0.069 \\

& CHIEF & Swin-tiny & 0.818$\pm$0.011 & 0.822$\pm$0.008 & 0.735$\pm$0.003 & 0.359$\pm$0.201 & 0.517$\pm$0.419 & 0.355$\pm$0.210 \\

& Lunit-DINO & ViT-small & 0.816$\pm$0.004 & 0.829$\pm$0.015 & 0.763$\pm$0.010 & 0.344$\pm$0.277 & 0.353$\pm$0.413 & 0.482$\pm$0.155 \\
& PLIP & \trinh{ViT-base/32} & 0.819$\pm$0.009 & 0.820$\pm$0.008 & 0.752$\pm$0.007 & 0.392$\pm$0.326 & 0.465$\pm$0.405 & 0.470$\pm$0.161 \\
& MAE-TCGA-Unlabeled & ViT-base & 0.806$\pm$0.017 & 0.826$\pm$0.009 & 0.749$\pm$0.005 & 0.377$\pm$0.071 & 0.698$\pm$0.064 & 0.466$\pm$0.074 \\
& SupMAE$^{\ddagger}$ & ViT-base & 0.755$\pm$0.015 & 0.800$\pm$0.026 & 0.671$\pm$0.029 & 0.624$\pm$0.009 & 0.735$\pm$0.012 & 0.592$\pm$0.012 \\

& UNI-DINOv2 & ViT-large & 0.837$\pm$0.009 & 0.835$\pm$0.009 & 0.776$\pm$0.008 & 0.364$\pm$0.215 & 0.624$\pm$0.342 & 0.445$\pm$0.107 \\

& GPFM & ViT-large & 0.794$\pm$0.016 & 0.835$\pm$0.003 & 0.751$\pm$0.007 & 0.421$\pm$0.211 & 0.701$\pm$0.125 & 0.393$\pm$0.040 \\

& Virchow & ViT-huge & 0.848$\pm$0.004 & 0.844$\pm$0.006 & 0.787$\pm$0.002 & 0.300$\pm$0.244 & 0.476$\pm$0.406 & 0.266$\pm$0.282 \\

\midrule
\multirow{4}{3cm}{\textit{Pretraining methods using our pseudo-labeled dataset: TCGA-MF-Pseudo}} & ViT-TCGA-MF-Pseudo & ViT-base & 0.825$\pm$0.005 & 0.823$\pm$0.006 & 0.766$\pm$0.004 & 0.824$\pm$0.001 & 0.821$\pm$0.005 & 0.766$\pm$0.004 \\

& MAE-TCGA-MF-Pseudo & ViT-base & 0.827$\pm$0.006 & 0.840$\pm$0.005 & 0.746$\pm$0.007 & 0.323$\pm$0.100 & 0.743$\pm$0.045 & 0.548$\pm$0.021 \\

& CMAE & ViT-base & 0.838$\pm$0.005 & 0.824$\pm$0.008 & 0.770$\pm$0.009 & 0.453$\pm$0.059 & 0.616$\pm$0.036 & 0.546$\pm$0.055 \\

& MiTHras (Ours) & ViT-base & \textbf{0.864$\pm$0.004} & \textbf{0.855$\pm$0.001} & \textbf{0.789$\pm$0.011} & \textbf{0.836$\pm$0.003} & \textbf{0.848$\pm$0.005} & \textbf{0.775$\pm$0.002} \\

\bottomrule
\end{tabular}
\end{adjustbox}
\end{center}
\end{table*}

\begin{figure}[!t]
\centering
\includegraphics[width=0.48\textwidth]{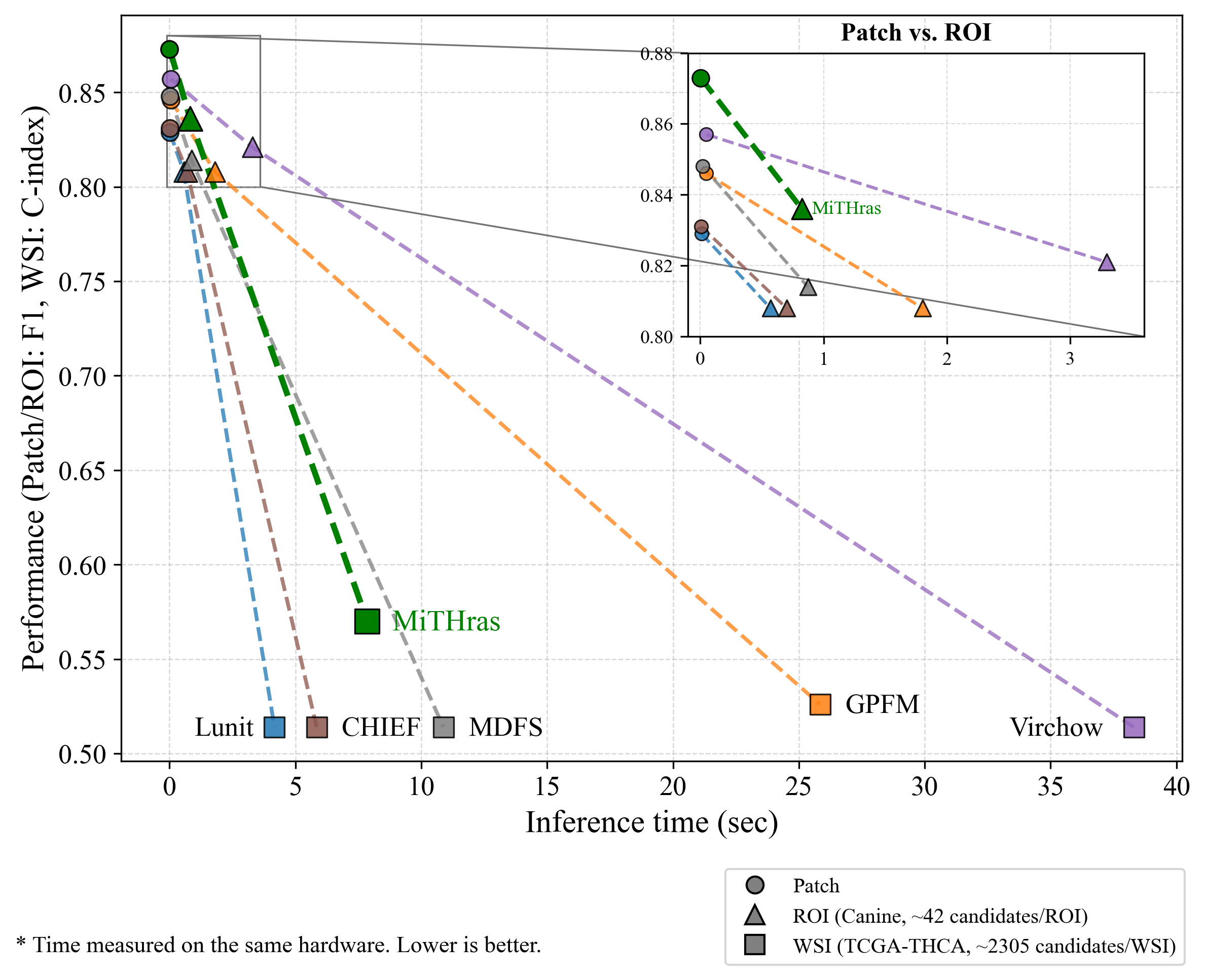}
\caption{\trinh{Performance and inference time at three evaluation scales. Circles denote classification of a single $128\times128$ patch; triangles denote detection in a Canine ROI (approximately 42 candidates); squares denote whole-slide inference on TCGA-THCA (approximately 2,305 candidates). The horizontal axis shows inference time, and the vertical axis shows F$_1$ for patch and ROI evaluation and the Cox C-index for WSI evaluation. Performance values are comparable within each
evaluation scale. Dashed lines connect the same model across scales; the inset enlarges the patch and ROI region. Times were measured on one NVIDIA RTX A6000 GPU with mixed precision.}}
\label{fig:mitosis_cls_line}
\end{figure}

\trinh{MiTHras attained the highest mean F$_1$ across all three test sets (Table~\ref{table:results_mitosis_cls_f1}). Under full fine-tuning, its margins over the strongest baseline, Virchow, were $0.016$ on MIDOGpp, $0.011$ on Canine, and $0.002$ on TCGA-MF-Test, though the latter difference was small relative to the reported fold variability.}
\trinh{This gap widened under linear probing. MiTHras achieved F$_1$ scores of $0.836$, $0.848$, and $0.775$ on MIDOGpp, Canine, and TCGA-MF-Test, respectively, with reductions of $0.007$--$0.028$ from full fine-tuning. On MIDOGpp, the best general-purpose or pathology foundation encoder reached $0.421$, while all encoders exceeding $0.45$ were pretrained on labeled or pseudo-labeled MF data. These results indicate that MiTHras features are more readily applicable without complex tuning.}

\trinh{The effect of the pretraining corpus varies with the objective and evaluation protocol. Supervised ViT pretraining on TCGA-MF-Pseudo produced substantially higher linear-probing F$_1$ than supervised ImageNet pretraining on all three datasets. For MAE, TCGA-MF-Pseudo improved linear probing over ImageNet universally, but fine-tuning results were mixed, with lower F$_1$ on MIDOGpp and TCGA-MF-Test and identical F$_1$ on Canine. MiTHras exceeded all three MAE baselines under both protocols, supporting the contribution of task-specific SCL beyond masked reconstruction.}
Among pathology foundation models, Virchow attained the highest fine-tuned F$_1$ on all three datasets, outperforming smaller encoders such as CTransPath-MoCov3 \cite{wang2022transformer} and CHIEF \cite{wang2024pathology}.

\subsection{Mitotic figure detection results} \label{section:results_mitosis_detection}

\begin{table}[h]
\begin{center}
\caption{\trinh{Mitotic figure detection results (F$_1$, mean $\pm$ standard deviation over three folds). For each stage-based model, the detection/classification threshold pair is selected to maximize F$_1$ across the three MIDOG22-CV folds. This pair is applied to MIDOGpp, Canine, and TCGA-MF-Test.  YOLOv10 uses its own inference confidence threshold. The Efficient-UNet-b0 segmentation stage is trained using the three-fold MDFS protocol and shared across all second-stage classifier comparisons. $^{\ddagger}$SupMAE is pretrained on the labeled MIDOG22 images. Best value per column in bold.}}
\label{table:results_mitosis_detection_foldbest_f1}
\setlength{\tabcolsep}{2pt}
\renewcommand{\arraystretch}{1}
\begin{adjustbox}{width=\columnwidth}
\begin{tabular}{l|c|c|c|c}
\toprule
Pretraining Method & Network & \multicolumn{1}{c|}{MIDOGpp} & \multicolumn{1}{c|}{Canine} & \multicolumn{1}{c}{TCGA-MF-Test} \\
\midrule
\multicolumn{5}{l}{\textit{Detection baselines}} \\
\noalign{\smallskip}
\trinh{Efficient-UNet-b0} (only $f^{\text{seg}}$)  & Efficient-UNet-b0 & 0.782$\pm$0.000 & 0.783$\pm$0.000 & 0.571$\pm$0.000 \\
YOLOv10 & YOLOv10 & 0.760$\pm$0.021 & 0.791$\pm$0.000 & 0.562$\pm$0.000 \\
\midrule
\multicolumn{5}{l}{\textit{\trinh{Natural-image pretraining}}} \\
\noalign{\smallskip}
MDFS & EfficientNet-b7 & 0.810$\pm$0.002 & 0.785$\pm$0.003 & 0.591$\pm$0.003 \\
ViT-ImageNet (Supervised) & ViT-base & 0.810$\pm$0.002 & 0.789$\pm$0.006 & 0.582$\pm$0.006 \\
MAE-ImageNet & ViT-base & 0.810$\pm$0.006 & 0.791$\pm$0.007 & 0.584$\pm$0.003 \\
DINOv2 & ViT-base & 0.813$\pm$0.001 & 0.795$\pm$0.002 & 0.592$\pm$0.002 \\
CLIP & ViT-base & 0.807$\pm$0.004 & 0.791$\pm$0.001 & 0.590$\pm$0.003 \\
\midrule
\multicolumn{5}{l}{\textit{Histopathology pretraining models}} \\
\noalign{\smallskip}
CTransPath-MoCov3 & Swin-tiny & 0.789$\pm$0.007 & 0.791$\pm$0.005 & 0.575$\pm$0.001 \\
CHIEF & Swin-tiny & 0.797$\pm$0.006 & 0.791$\pm$0.005 & 0.572$\pm$0.002 \\
Lunit-DINO & ViT-small & 0.802$\pm$0.006 & 0.789$\pm$0.006 & 0.574$\pm$0.005 \\
PLIP & \trinh{ViT-base/32} & 0.802$\pm$0.003 & 0.784$\pm$0.007 & 0.577$\pm$0.003 \\

MAE-TCGA-Unlabeled & ViT-base & 0.799$\pm$0.006 & 0.789$\pm$0.002 & 0.568$\pm$0.002 \\
UNI-DINOv2 & ViT-large & 0.814$\pm$0.007 & 0.795$\pm$0.002 & 0.587$\pm$0.005 \\
GPFM & ViT-large & 0.784$\pm$0.009 & \textbf{0.798$\pm$0.003} & 0.575$\pm$0.004 \\
Virchow & ViT-huge & 0.816$\pm$0.002 & 0.796$\pm$0.002 & 0.589$\pm$0.001 \\
SupMAE$^{\ddagger}$ & ViT-base & 0.767$\pm$0.005 & 0.774$\pm$0.013 & 0.540$\pm$0.007 \\
\midrule
\multicolumn{5}{l}{\textit{Pretraining methods using our pseudo-labeled dataset: TCGA-MF-Pseudo}} \\
\noalign{\smallskip}
ViT-TCGA-MF-Pseudo & ViT-base & 0.809$\pm$0.003 & 0.791$\pm$0.002 & 0.586$\pm$0.003 \\
MAE-TCGA-MF-Pseudo & ViT-base & 0.804$\pm$0.003 & 0.791$\pm$0.003 & 0.569$\pm$0.004 \\
CMAE & ViT-base & 0.810$\pm$0.003 & 0.785$\pm$0.005 & 0.582$\pm$0.006 \\
MiTHras (Ours) & ViT-base & \textbf{0.829$\pm$0.002} & 0.797$\pm$0.006 & \textbf{0.599$\pm$0.005} \\
\bottomrule
\end{tabular}
\end{adjustbox}
\end{center}
\end{table}
\trinh{Table~\ref{table:results_mitosis_detection_foldbest_f1} compares detection performance. MiTHras obtained the highest mean F$_1$ on MIDOGpp ($0.829$) and TCGA-MF-Test ($0.599$), and was comparable to the best model (GPFM) on Canine (0.797 vs. 0.798). Differences among second-stage classifiers were smaller than in the classification experiment, as the shared first-stage detector constrains candidate recall. Nevertheless, most two-stage models outperformed the standalone segmentation model $f^{\text{seg}}$, confirming that the second stage contributes by separating true MFs from mimickers.}
\trinh{Among the two-stage methods, SupMAE, pretrained on the smaller labeled MIDOG22 corpus, yielded the lowest mean F$_1$ on all three test sets. Foundation-model rankings varied by dataset: Virchow led this group on MIDOGpp and TCGA-MF-Test, whereas GPFM led on Canine.}

\subsection{Mitotic count for survival analysis}
\label{sec:survival-results}

\trinh{We examined whether automated MF counts align with diagnostic grade and provide prognostic value. Table~\ref{table:survival} reports cohort-specific survival results and average second-stage inference time across TCGA-THCA and TCGA-GBMLGG. Figure~\ref{fig:mitosis_cls_line} illustrates inference timing for TCGA-THCA.}

\begin{figure}[!t]
\centering
\includegraphics[width=\columnwidth]{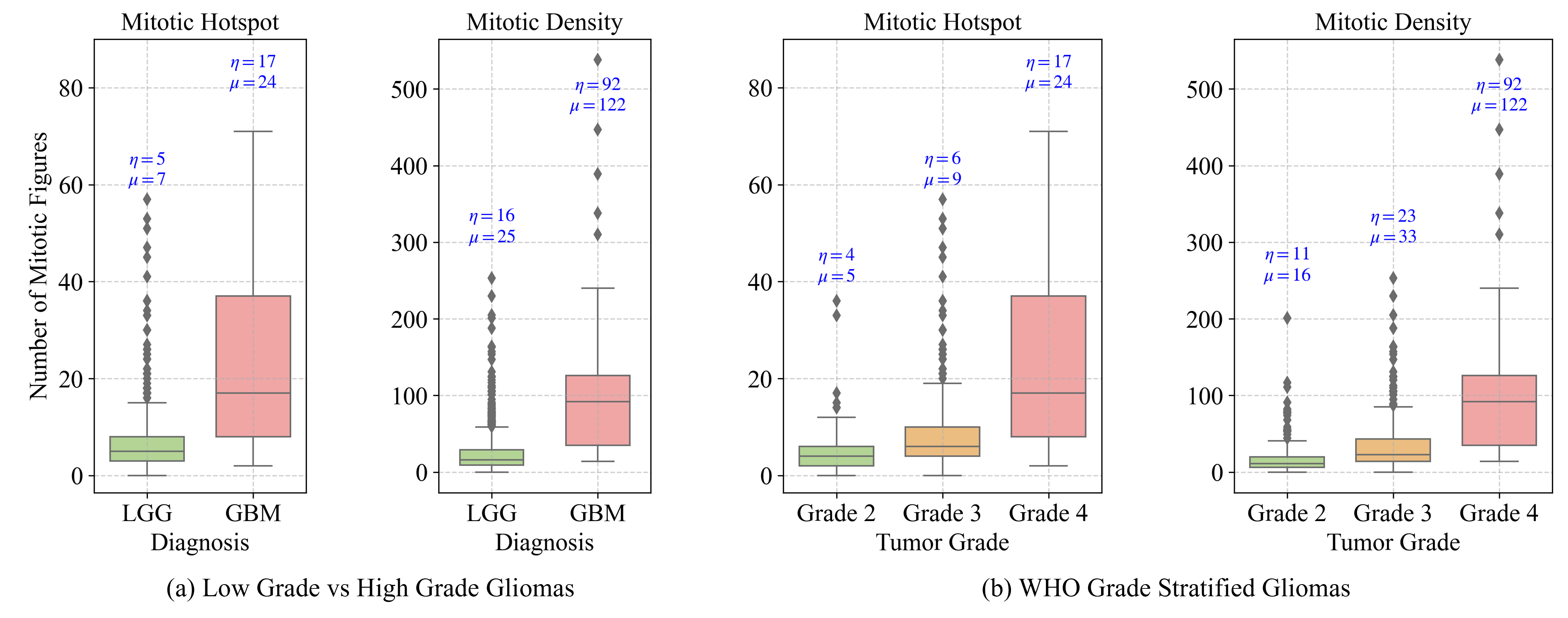}
\caption{Boxplots of mitotic counts across different diagnoses in the TCGA-GBMLGG cohorts. Median ($\eta$) and mean ($\mu$) values are indicated. The mitotic count is produced by MiTHras using mitotic hotspot (2 mm$^2$) and mitotic density (20 mm$^2$) methods.}
\label{fig:box_mitosis_count_diagnosis}
\end{figure}

\noindent\textbf{Correlation with Diagnostic Subtypes.} 
We compared mitotic counts produced by MiTHras under the two counting strategies, mitotic hotspot (2~mm$^2$) and mitotic density (20~mm$^2$), for two diagnostic tasks in \trinh{TCGA-GBMLGG} (Figure~\ref{fig:box_mitosis_count_diagnosis}): the lower-grade glioma (LGG) and glioblastoma (GBM) cohorts, and grade 2 versus 3 versus 4.
Median mitotic counts increased with tumor grade under both strategies (median hotspot count 5 in LGG versus 17 in GBM; 4, 6 and 17 across grades 2--4).
The two strategies diverged mainly on the intermediate grades: hotspot counts for grades 2 and 3 are nearly indistinguishable (medians 4 and 6), whereas density shows a larger difference in group medians (11 and 23).

\noindent\textbf{Survival Prediction.}
\trinh{Table~\ref{table:survival} reports the apparent C-index and 95\% bootstrap confidence interval for each count-based Cox model. On TCGA-GBMLGG, MiTHras obtained C-indices of $0.627$ (hotspot) and $0.671$ (density). On TCGA-THCA, all point estimates fell below $0.5$, with MiTHras yielding $0.446$ (95\% CI $0.313$--$0.607$) for hotspot and $0.427$ ($0.287$--$0.593$) for density. Given only 11 events across 395 patients (406 WSIs), these results are reported descriptively.}
\trinh{These C-indices evaluate mitotic count as a single morphological predictor. Samples with identical counts receive identical predicted risks, and these ties receive half credit in Harrell's C-index. Consequently, differences between methods reflect variations in detection coverage as well as actual biological variations across slides. We therefore present these cohort-level associations as an exploratory assessment of automated mitotic counting.}

\begin{table*}[!t]
\begin{center}
\caption{
\trinh{C-index for survival analysis on TCGA-THCA and TCGA-GBMLGG.
Time is the average wall-clock inference time of the second-stage classifier per WSI, measured on a single NVIDIA RTX A6000 GPU with automatic mixed precision. Relative time is this time normalized to the ViT-base configuration ($1.00\times$). All two-stage methods share the same Efficient-UNet-b0 segmentation stage. The segmentation-only baseline performs no classification. Best value per column is shown in bold for TCGA-GBMLGG only. No values are highlighted for TCGA-THCA, which is reported descriptively because only 11 deaths are recorded.}}

\label{table:survival}
\setlength{\tabcolsep}{6pt} 
\renewcommand{\arraystretch}{1} 
\begin{adjustbox}{width=\textwidth}
\begin{tabular}{ll|c|cc|cc|cc}
\toprule
\multirow{4}{*}{Pretraining Method} & & \multirow{4}{*}{Network ($f^{\text{cls}}$)} & \multicolumn{2}{c|}{TCGA-THCA} & \multicolumn{2}{c|}{TCGA-GBMLGG} & \multicolumn{2}{c}{Classifier inference}\\
\noalign{\smallskip}
\cline{3-9}
\noalign{\smallskip}

& & & C-index  & C-index& C-index  & C-index  & Time  & Relative time\\
& &  &Mitotic hotspot (2 mm$^2$) & Mitotic density (20 mm$^2$) & Mitotic hotspot (2 mm$^2$)  & Mitotic density (20 mm$^2$) & (second)  &  \\
 \midrule

\textit{Single stage models} & \trinh{Efficient-UNet-b0} (only $f^{\text{seg}}$) & - & 0.437 (0.405, 0.485) & 0.438 (0.405, 0.490) & 0.540 (0.508, 0.574) & 0.541 (0.509, 0.575) & - & - \\
 \midrule
\multirow{4}{3cm}{\textit{\trinh{Natural-image pretraining}}} & MDFS & EfficientNet-b7 & 0.417 (0.381, 0.475) & 0.417 (0.381, 0.475) & 0.541 (0.504, 0.575) & 0.543 (0.507, 0.577) & 9.17 & 1.35 $\times$ \\
& ViT-ImageNet & ViT-base & 0.435 (0.358, 0.564) & 0.433 (0.357, 0.562) & 0.528 (0.488, 0.566) & 0.533 (0.493, 0.572) & 6.80 & 1.00 $\times$ \\
& MAE-ImageNet & ViT-base & 0.437 (0.358, 0.565) & 0.434 (0.356, 0.562) & 0.527 (0.487, 0.566) & 0.533 (0.492, 0.571) & 6.80 & 1.00 $\times$ \\
& CLIP & ViT-base & 0.417 (0.381, 0.475) & 0.417 (0.381, 0.475) & 0.539 (0.503, 0.573) & 0.542 (0.506, 0.575) & 6.80 & 1.00 $\times$ \\
\midrule
\multirow{9}{3cm}{\textit{Histopathology pretraining models}} & CTransPath-MoCov3 & Swin-tiny & 0.416 (0.381, 0.473) & 0.415 (0.381, 0.468) & 0.540 (0.504, 0.574) & 0.542 (0.506, 0.575) & 5.77 & 0.85 $\times$ \\
& CHIEF & Swin-tiny & 0.417 (0.381, 0.475) & 0.418 (0.381, 0.480) & 0.540 (0.503, 0.574) & 0.541 (0.505, 0.574) & 5.77 & 0.85 $\times$ \\
& Lunit-DINO & ViT-small & 0.417 (0.381, 0.475) & 0.417 (0.381, 0.478) & 0.539 (0.503, 0.572) & 0.541 (0.504, 0.573) & 3.71 & 0.55 $\times$ \\
& PLIP & \trinh{ViT-base/32} & 0.419 (0.382, 0.480) & 0.419 (0.382, 0.482) & 0.538 (0.502, 0.572) & 0.541 (0.505, 0.574) & 3.02 & 0.44 $\times$ \\

& MAE-TCGA-Unlabeled & ViT-base & 0.415 (0.381, 0.469) & 0.416 (0.381, 0.471) & 0.539 (0.503, 0.573) & 0.541 (0.505, 0.574) &   6.80 & 1.00 $\times$\\
& SupMAE & ViT-base & 0.416 (0.381, 0.472) & 0.416 (0.381, 0.473) & 0.539 (0.503, 0.573) & 0.541 (0.505, 0.574) & 6.80 & 1.00 $\times$ \\

& UNI-DINOv2 & ViT-large & 0.436 (0.357, 0.564) & 0.436 (0.357, 0.564) & 0.527 (0.487, 0.565) & 0.534 (0.493, 0.573) & 21.94 & 3.23 $\times$ \\
& GPFM & ViT-large & 0.417 (0.381, 0.478) & 0.417 (0.381, 0.478) & 0.540 (0.504, 0.574) & 0.541 (0.505, 0.574) & 21.94 & 3.23 $\times$ \\
& Virchow & ViT-huge & 0.417 (0.381, 0.474) & 0.417 (0.381, 0.477) & 0.541 (0.504, 0.574) & 0.544 (0.507, 0.577) & 32.60 & 4.79 $\times$ \\

\midrule
\multirow{4}{3cm}{\textit{Pretraining methods using our pseudo-labeled dataset: TCGA-MF-Pseudo}} & ViT-TCGA-MF-Pseudo & ViT-base & 0.417 (0.381, 0.478) & 0.419 (0.381, 0.481) & 0.540 (0.503, 0.574) & 0.542 (0.505, 0.575) & 6.80 & 1.00 $\times$ \\
& MAE-TCGA-MF-Pseudo & ViT-base & 0.433 (0.357, 0.561) & 0.435 (0.357, 0.563) & 0.529 (0.489, 0.566) & 0.533 (0.493, 0.571) & 6.80 & 1.00 $\times$ \\
& CMAE & ViT-base & 0.433 (0.357, 0.559) & 0.432 (0.357, 0.558) & 0.528 (0.488, 0.567) & 0.531 (0.491, 0.570) & 6.80 & 1.00 $\times$ \\
& MiTHras & ViT-base & 0.446 (0.313, 0.607) & 0.427 (0.287, 0.593) & \textbf{0.627 (0.580, 0.671)} & \textbf{0.671 (0.630, 0.713)} & 6.80 & 1.00 $\times$ \\

\bottomrule
\end{tabular}
\end{adjustbox}
\end{center}
\end{table*}

\subsection{Mitotic figure subtype classification results}
\label{sec:subtype-results} 

\begin{table}[h]
\begin{center}
\caption{Mitotic figure subtype classification results across AMi-Br-TUPAC and AMi-Morph-TUPAC. \trinh{All values are F$_1$, reported as mean $\pm$ standard deviation over three folds. $^{\ddagger}$SupMAE is pretrained on the labeled MIDOG22 images. Best value per column in bold.}}
\label{table:results_mitosis_subtype}
\setlength{\tabcolsep}{6pt} 
\renewcommand{\arraystretch}{0.9} 
\begin{adjustbox}{width=\columnwidth}
\begin{tabular}{l|c|c|c}
\toprule
\multirow{2}{*}{Pretraining Method} & \multirow{2}{*}{Network} & \multicolumn{1}{c|}{AMi-Br-TUPAC} & \multicolumn{1}{c}{AMi-Morph-TUPAC} \\
& & (typical vs. atypical) & (8 subtypes) \\
\midrule
\multicolumn{4}{l}{\textit{\trinh{Natural-image pretraining}}} \\
\noalign{\smallskip}
Bertram et al.\ \cite{bertram2025histologic} & EfficientNet-V2-S & 0.577$\pm$0.019 & 0.308$\pm$0.027 \\
MDFS & EfficientNet-b7 & 0.583$\pm$0.038 & 0.410$\pm$0.039 \\
ViT-ImageNet (Supervised) & ViT-base & 0.623$\pm$0.005 & 0.439$\pm$0.012 \\
MAE-ImageNet & ViT-base & 0.456$\pm$0.251 & 0.413$\pm$0.030 \\
DINOv2 & ViT-base & 0.654$\pm$0.011 & 0.445$\pm$0.032 \\
CLIP & ViT-base & 0.643$\pm$0.018 & 0.445$\pm$0.020 \\

\midrule
\multicolumn{4}{l}{\textit{Histopathology pretraining models}} \\
\noalign{\smallskip}
CTransPath-MoCov3 & Swin-tiny & 0.228$\pm$0.203 & 0.319$\pm$0.050 \\
CHIEF & Swin-tiny & 0.542$\pm$0.007 & 0.305$\pm$0.018 \\
PLIP & \trinh{ViT-base/32} & 0.642$\pm$0.003 & 0.434$\pm$0.026 \\
Lunit-DINO & ViT-small & 0.575$\pm$0.015 & 0.351$\pm$0.029 \\
MAE-TCGA-Unlabeled & ViT-base & 0.246$\pm$0.062 & 0.303$\pm$0.004 \\
SupMAE$^{\ddagger}$ & ViT-base & 0.505$\pm$0.035 & 0.376$\pm$0.034 \\
UNI-DINOv2 & ViT-large & 0.568$\pm$0.025 & 0.357$\pm$0.026 \\
GPFM & ViT-large & 0.653$\pm$0.013 & 0.452$\pm$0.034 \\
Virchow & ViT-huge & 0.570$\pm$0.011 & 0.329$\pm$0.020 \\

\midrule
\multicolumn{4}{l}{\textit{Pretraining methods using our pseudo-labeled dataset: TCGA-MF-Pseudo}} \\
\noalign{\smallskip}
ViT-TCGA-MF-Pseudo & ViT-base & 0.535$\pm$0.035 & 0.252$\pm$0.007 \\
MAE-TCGA-MF-Pseudo & ViT-base & 0.100$\pm$0.058 & 0.225$\pm$0.086 \\
CMAE & ViT-base & 0.597$\pm$0.014 & 0.362$\pm$0.006 \\
MiTHras (Ours) & ViT-base & \textbf{0.700$\pm$0.005} & \textbf{0.483$\pm$0.006} \\

\bottomrule
\end{tabular}
\end{adjustbox}
\end{center}
\end{table}

Table~\ref{table:results_mitosis_subtype} presents MF subtype classification results. \trinh{MiTHras achieved the highest mean F$_1$ on both benchmarks: $0.700$ on AMi-Br-TUPAC and $0.483$ on AMi-Morph-TUPAC, outperforming the strongest baselines by $0.046$ and $0.031$, respectively. 
MiTHras also outperformed SupMAE and the ViT-, MAE-, and CMAE-based models pretrained on TCGA-MF-Pseudo on both benchmarks. Several pathology foundation models, including CTransPath-MoCov3, CHIEF, UNI-DINOv2, and Virchow, fell below the supervised ViT-ImageNet baseline. Eight-class morphology remains challenging, yielding a maximum macro F$_1$ of $0.483$, in which the metric differs from the positive-class F$_1$ used for AMi-Br.}

\subsection{Ablation study}
\label{sec:ablation} 

\trinh{Table~\ref{table:results_mitosis_cls_Ablation} shows the effect of loss components, $\beta$, backbone, and pretraining corpus. Regarding the loss components, masked reconstruction ($\mathcal{L}_{MIM}$) alone performed poorly. Incorporating image-level SCL ($\mathcal{L}_{SCL}^{img}$) substantially improved performance, increasing F$_1$ across all eight evaluations and raising the descriptive mean from $0.544$ to $0.766$, with the largest gains observed in subtype classification. 
Notably, $\mathcal{L}_{SCL}^{img}$ alone produced variable results across datasets. These results indicate that neither term alone is sufficient for consistent and reliable performance.}

\trinh{Incorporating token-level SCL ($\mathcal{L}_{SCL}^{tok}$) produced task-dependent trade-offs. Relative to using $\mathcal{L}_{MIM}$ and $\mathcal{L}_{SCL}^{img}$ ($\beta=1.0$), the main configuration ($\beta=0.75$) enhanced the overall mean from $0.766$ to $0.769$ but yielded mixed task-level results: AMi-Br F$_1$ improved from $0.667$ to $0.700$, while AMi-Morph F$_1$ dropped from $0.494$ to $0.483$. Fine-tuned MF-versus-mimicker F$_1$ decreased marginally by $0.001-0.010$ across MIDOGpp, Canine, and TCGA-MF-Test. Linear probing showed gains for Canine ($+0.009$) and TCGA-MF-Test ($+0.016$), but a drop for MIDOGpp ($-0.011$). Thus, token-level supervision does not consistently improve fine morphological discrimination.
}
\trinh{Moreover, excluding $\mathcal{L}_{MIM}$ ($0.75\mathcal{L}_{SCL}^{img}+0.25\mathcal{L}_{SCL}^{tok}$) had divergent effects under fine-tuning and linear probing. It lowered F$_1$ on all three fine-tuned MF-versus-mimicker benchmarks, yet produced mixed linear-probing results: MIDOGpp and TCGA-MF-Test increased, while Canine decreased. AMi-Br remained at $0.700$, but AMi-Morph decreased from $0.483$ to $0.475$. 
}

\trinh{Backbone scale and pretraining corpus also affect transferability. With ViT-tiny, $\beta=0.75$ improved all three linear-probing and both subtype results over $\beta=1$, while lowering all three fine-tuned MF-versus-mimicker results. ViT-tiny approached ViT-base in binary classification, though its AMi-Morph F$_1$ remained lower. For ViT-base at $\beta=0.75$, pretraining on TCGA-MF-Pseudo produced an eight-evaluation mean of $0.769$, surpassing $0.684$ obtained with labeled MIDOG22, highlighting the benefits of the larger scale and diversity of the pseudo-labeled corpus.}

\trinh{To assess the impact of the tumor filter, we compared IMPaSh and CONCH. IMPaSh retained $14.7\%$ of the sampled patches, compared with $27.0\%$ for CONCH. In the pathologist-reviewed disagreement strata, pooled tumor precision was $87.7$--$95.4\%$ for patches retained only by IMPaSh and $87.7$--$92.9\%$ for patches retained only by CONCH (Table~\ref{tab:tumor-review}). 
Performance varied across cohorts, with neither filter consistently achieving higher precision. In downstream evaluation at $\beta=0.75$, replacing IMPaSh with CONCH decreased the eight-evaluation mean F$_1$ from $0.769$ to $0.765$. At $\beta=1$, the corresponding means were comparable ($0.766$ vs. $0.767$; Table~\ref{table:results_mitosis_cls_Ablation}). Overall, although some individual subtype variations were larger, the aggregate differences remained minor.
}

\begin{table*}[t]
\begin{center}
\caption{\trinh{Results of ablation experiments. Unless stated otherwise, rows use ViT-base pretrained on TCGA-MF-Pseudo with the IMPaSh tumor filter. Results are mean F$_1$ $\pm$ standard deviation over three folds. Avg. is a descriptive, unweighted mean of the eight displayed evaluations. The shaded row is the main configuration. Bold marks the highest displayed mean in each column.}}

\label{table:results_mitosis_cls_Ablation}
\setlength{\tabcolsep}{3pt}
\renewcommand{\arraystretch}{0.95}
\begin{adjustbox}{width=\textwidth}
\trinh{\begin{tabular}{l|ccc|ccc|cc|c}
\toprule
\multirow{2}{*}{Configuration} & \multicolumn{3}{c|}{Fine-tuning} & \multicolumn{3}{c|}{Linear probing} & \multicolumn{2}{c|}{Fine-tuning} & \multirow{2}{*}{Avg.} \\
& MIDOGpp & Canine & TCGA-MF-Test & MIDOGpp & Canine & TCGA-MF-Test & AMi-Br & AMi-Morph & \\
\midrule
\multicolumn{10}{l}{\textit{Loss components}} \\
$\mathcal{L}_{\text{MIM}}$ only (MAE) & 0.827$\pm$0.006 & 0.840$\pm$0.005 & 0.746$\pm$0.007 & 0.323$\pm$0.100 & 0.743$\pm$0.045 & 0.548$\pm$0.021 & 0.100$\pm$0.058 & 0.225$\pm$0.086 & 0.544 \\
$\mathcal{L}^{\text{img}}_{\text{SCL}}$ only & 0.857$\pm$0.008 & 0.854$\pm$0.001 & 0.788$\pm$0.013 & 0.833$\pm$0.001 & 0.843$\pm$0.002 & \textbf{0.813$\pm$0.002} & 0.677$\pm$0.015 & 0.458$\pm$0.012 & 0.765 \\
$0.75\mathcal{L}^{\text{img}}_{\text{SCL}}+0.25\mathcal{L}^{\text{tok}}_{\text{SCL}}$ (no MIM) & 0.726$\pm$0.033 & 0.694$\pm$0.026 & 0.648$\pm$0.021 & 0.839$\pm$0.002 & 0.843$\pm$0.004 & 0.804$\pm$0.003 & \textbf{0.700$\pm$0.005} & 0.475$\pm$0.030 & 0.716 \\
$\mathcal{L}_{\text{MIM}} + \mathcal{L}^{\text{tok}}_{\text{SCL}}$ ($\beta=0$) & 0.862$\pm$0.005 & 0.859$\pm$0.006 & 0.781$\pm$0.011 & 0.801$\pm$0.002 & 0.833$\pm$0.001 & 0.750$\pm$0.002 & 0.676$\pm$0.010 & 0.475$\pm$0.023 & 0.755 \\
$\mathcal{L}_{\text{MIM}} + \mathcal{L}^{\text{img}}_{\text{SCL}}$ ($\beta=1$) & \textbf{0.865$\pm$0.009} & 0.857$\pm$0.005 & \textbf{0.799$\pm$0.009} & \textbf{0.847$\pm$0.004} & 0.839$\pm$0.002 & 0.759$\pm$0.001 & 0.667$\pm$0.016 & 0.494$\pm$0.020 & 0.766 \\
\midrule
\multicolumn{10}{l}{\textit{Sensitivity to $\beta$ (all three terms)}} \\
$\beta=0.25$ & 0.862$\pm$0.004 & 0.856$\pm$0.002 & 0.787$\pm$0.004 & 0.839$\pm$0.004 & 0.830$\pm$0.004 & 0.775$\pm$0.005 & 0.690$\pm$0.014 & 0.457$\pm$0.023 & 0.762 \\
$\beta=0.50$ & 0.861$\pm$0.004 & 0.858$\pm$0.002 & 0.793$\pm$0.006 & 0.842$\pm$0.004 & 0.835$\pm$0.005 & 0.781$\pm$0.001 & 0.649$\pm$0.010 & 0.468$\pm$0.003 & 0.761 \\
\rowcolor{gray!15}
$\beta=0.75$ (MiTHras) & 0.864$\pm$0.004 & 0.855$\pm$0.001 & 0.789$\pm$0.011 & 0.836$\pm$0.003 & 0.848$\pm$0.005 & 0.775$\pm$0.002 & \textbf{0.700$\pm$0.005} & 0.483$\pm$0.006 & \textbf{0.769} \\
\midrule
\multicolumn{10}{l}{\textit{\trinh{ViT-tiny backbone}}} \\
$\beta=1$ & 0.862$\pm$0.004 & \textbf{0.860$\pm$0.002} & 0.792$\pm$0.003 & 0.838$\pm$0.003 & 0.843$\pm$0.002 & 0.765$\pm$0.000 & 0.659$\pm$0.022 & 0.343$\pm$0.005 & 0.745 \\
$\beta=0.75$ & 0.858$\pm$0.002 & 0.856$\pm$0.001 & 0.787$\pm$0.004 & 0.843$\pm$0.001 & 0.846$\pm$0.003 & 0.782$\pm$0.002 & 0.670$\pm$0.012 & 0.394$\pm$0.055 & 0.754 \\
\midrule
\multicolumn{10}{l}{\textit{Pretraining methods using labeled dataset: MIDOG22}} \\
$\beta=1$ & 0.781$\pm$0.020 & 0.812$\pm$0.002 & 0.711$\pm$0.008 & 0.764$\pm$0.016 & 0.794$\pm$0.020 & 0.683$\pm$0.024 & 0.511$\pm$0.032 & 0.358$\pm$0.010 & 0.677 \\
$\beta=0.75$ & 0.781$\pm$0.011 & 0.816$\pm$0.008 & 0.721$\pm$0.009 & 0.765$\pm$0.011 & 0.790$\pm$0.020 & 0.687$\pm$0.021 & 0.553$\pm$0.023 & 0.362$\pm$0.028 & 0.684 \\
\midrule
\multicolumn{10}{l}{\textit{\trinh{TCGA-MF-Pseudo constructed with CONCH instead of IMPaSh}}} \\
$\beta=1$ & \textbf{0.865$\pm$0.006} & 0.856$\pm$0.006 & 0.788$\pm$0.004 & 0.839$\pm$0.002 & 0.848$\pm$0.003 & 0.777$\pm$0.002 & 0.686$\pm$0.016 & 0.473$\pm$0.048 & 0.767 \\
$\beta=0.75$ & 0.857$\pm$0.001 & 0.856$\pm$0.006 & 0.790$\pm$0.006 & 0.827$\pm$0.004 & \textbf{0.851$\pm$0.002} & 0.790$\pm$0.003 & 0.679$\pm$0.016 & 0.473$\pm$0.011 & 0.765 \\
\midrule
\multicolumn{10}{l}{\textit{\trinh{MiTHras with 165,514 interchanged pseudo labels}}} \\
$\beta=0.75$ & 0.856$\pm$0.005 & 0.847$\pm$0.013 & 0.784$\pm$0.008 & 0.839$\pm$0.002 & 0.836$\pm$0.004 & 0.768$\pm$0.003 & 0.683$\pm$0.006 & \textbf{0.496$\pm$0.023} & 0.764 \\
\bottomrule
\end{tabular}}
\end{adjustbox}
\end{center}
\end{table*}

\begin{table}[t]
\centering
\setlength{\tabcolsep}{8pt}
\caption{\trinh{Tumor-patch selection: pathologist-confirmed precision of IMPaSh and CONCH retention decisions, 25 patches per stratum per cohort. Ranges span ``unsure'' counted as incorrect or correct.}}
\label{tab:tumor-review}
\begin{adjustbox}{width=0.48\textwidth}
\trinh{\begin{tabular}{lccc}
\toprule
Cohort & Both & CONCH only & IMPaSh only \\
\midrule
Bladder (BLCA)        & 1.00 & 0.96--1.00 & 0.96--1.00 \\
Breast (BRCA)         & 1.00 & 0.88       & 0.92--1.00 \\
Colon (COLON\_DX)     & 0.96 & 0.92--0.96 & 0.92 \\
Esophagus (ESCA)      & 1.00 & 0.88--0.92 & 0.72--0.88 \\
Renal chromophobe (KICH) & 0.96 & 0.92    & 0.80--0.92 \\
Renal clear cell (KIRC)  & 1.00 & 1.00    & 0.88--1.00 \\
Renal papillary (KIRP)   & 1.00 & 0.92--1.00 & 0.84--0.96 \\
Liver (LIHC)          & 1.00 & 1.00       & 0.96--1.00 \\
Lung adeno. (LUAD)    & 0.96--1.00 & 0.56--0.72 & 0.88--0.96 \\
Lung squamous (LUSC)  & 1.00 & 0.72--0.76 & 1.00 \\
Ovary (OV)            & 1.00 & 0.88--0.92 & 0.76--0.92 \\
Pancreas (PAAD)       & 0.92 & 0.84--0.96 & 0.84 \\
Prostate (PRAD)       & 0.96--1.00 & 0.88--1.00 & 0.92--1.00 \\
Uterus (UCEC)         & 1.00 & 0.92--0.96 & 0.88--0.96 \\
\midrule
\textbf{Pooled} ($n=350$) & \textbf{0.983--0.989} & \textbf{0.877--0.929} & \textbf{0.877--0.954} \\
\bottomrule
\end{tabular}}
\end{adjustbox}
\end{table}

\subsection{\trinh{Pseudo-label quality}}
{\revisioncolor
\label{sec:pseudo-validity}
The blinded review of 420 pseudo-mitotic review records yielded pool-weighted precision of $94.9$--$98.1\%$ (unweighted: $92.9$--$97.1\%$; Table~\ref{tab:mf-review}). In the additional-figure assessment, 8 of the 199 crops ($4.02\%$; 95\% Wilson CI, $2.05$--$7.73\%$) contained another MF, indicating that crops centered on non-mitotic objects can still capture peripheral MFs, introducing errors into pseudo labels.
The bounding-box assessment included 445 reviews of confirmed MFs (390 pseudo-mitotic and 55 pseudo-mimicker). Overall, $81.6\%$ of reviewed MFs ($363/445$) fit within a box of at most $50\times50$ pixels. Specifically, $3.1\%$ of MFs fit within a box of 25 pixels, $78.4\%$ within 50 pixels, $14.2\%$ within 75 pixels, and $4.3\%$ exceeded 75 pixels. The fixed pseudo box therefore provides coarse spatial supervision for token-level learning.
}

\begin{table}[t]
\centering
\setlength{\tabcolsep}{7pt}
\caption{\trinh{Mitotic pseudo-label precision by cohort, 30 pseudo-mitotic crop reviews per cohort. Pooled value is weighted by cohort pool size.}}
\label{tab:mf-review}
\begin{adjustbox}{width=0.48\textwidth}
\trinh{\begin{tabular}{lrcc}
\toprule
Cohort & Pool size & Precision & 95\% CI (worst case) \\
\midrule
Breast (BRCA)            & 97,540  & 1.00        & 0.89--1.00 \\
Renal papillary (KIRP)   & 8,059   & 1.00        & 0.89--1.00 \\
Ovary (OV)               & 42,582  & 1.00        & 0.89--1.00 \\
Uterus (UCEC)            & 88,659  & 1.00        & 0.89--1.00 \\
Esophagus (ESCA)         & 33,332  & 0.97        & 0.83--0.99 \\
Colon (COLON\_DX)        & 105,805 & 0.93--0.97  & 0.79--0.98 \\
Lung squamous (LUSC)     & 100,249 & 0.93--0.97  & 0.79--0.98 \\
Pancreas (PAAD)          & 5,095   & 0.93--0.97  & 0.79--0.98 \\
Liver (LIHC)             & 55,433  & 0.90--1.00  & 0.74--0.97 \\
Bladder (BLCA)           & 68,396  & 0.90--0.97  & 0.74--0.97 \\
Lung adeno. (LUAD)       & 43,397  & 0.90--0.97  & 0.74--0.97 \\
Prostate (PRAD)          & 8,164   & 0.90--0.97  & 0.74--0.97 \\
Renal clear cell (KIRC)  & 1,780   & 0.83--0.97  & 0.66--0.93 \\
Renal chromophobe (KICH) & 2,340   & 0.80--0.87  & 0.63--0.90 \\
\midrule
\textbf{Pooled, pool-weighted} & \textbf{660,831} & \textbf{0.949--0.981} & N/A \\
\bottomrule
\end{tabular}}
\end{adjustbox}
\end{table}

\subsection{Effect of pseudo-label noise}
\label{sec:noise-analysis}

To evaluate sensitivity to pseudo-label errors, we repeated pretraining after swapping 165,514 labels ($9.224\%$ of the corpus: 82,757 pseudo-mitotic crops with classifier probabilities of 0.5--0.6 and 82,757 ranked pseudo-mimickers). The eight-evaluation mean F$_1$ decreased only from $0.769$ to $0.764$, with per-benchmark changes between $-0.017$ and $+0.013$ (Table~\ref{table:results_mitosis_cls_Ablation}).

\section{Discussion} \label{section:Discussions}
In this work, we proposed MiTHras, a task-specific hierarchical SCL framework for MF analysis. \trinh{MiTHras builds on ViT-base pretrained on 1.8 million images from 6,041 WSIs. In comparison to the ViT-large UNI-DINOv2 model, pretrained on 100 million images from 100,426 WSIs \cite{chen2024towards}, MiTHras achieved higher mean F$_1$ on the three classification and two subtype benchmarks, despite its smaller backbone and corpus. Furthermore, its second-stage classifier required lower inference time than the larger ViT-large and ViT-huge encoders (Table~\ref{table:survival}). These results underscore the value of aligning pretraining data and objectives with MF analysis.}

The tumor-filter comparison (Section~\ref{sec:ablation}) merits further explanation, given that IMPaSh--trained on labeled colorectal tissue patches--performed comparably to CONCH. We attribute this comparable efficacy to IMPaSh's capability in distinguishing non-tumor tissues rather than broad organ coverage. IMPaSh assigns each sub-patch to tumor or one of eight non-tumor categories. Aside from normal colon mucosa, the remaining categories (adipose, background, debris, lymphocytes, mucus, smooth muscle, and stroma) exhibit common morphological characteristics across diverse organs. Accordingly, filtering out these components may support tumor patch selection beyond the colorectal domain. CONCH, by comparison, learned broad pathology visual-language representations, without being explicitly optimized for tumor detection. Its broader organ coverage therefore need not translate into higher precision for this particular filtering task. Nevertheless, both filters operate as pre-processing steps to generate the pretraining MF and mimicker cropss. The small difference in eight-evaluation F$_1$ mean (IMPaSh: 0.769 vs. CONCH: 0.765) indicates limited aggregate sensitivity to this filtering choice.

\trinh{MiTHras integrates pseudo-label-guided local alignment with global class alignment and masked reconstruction. Local representation learning methods such as VICRegL \cite{bardes2022vicregl} use correspondences between augmented views, whereas MiTHras groups tokens by their mitotic or non-mitotic pseudo-labels. The binary token labels distinguish mitotic from non-mitotic regions--aiding typical-versus-atypical classification--they do not consistently enhance discrimination among eight morphology classes. This may be ascribable to the use of fixed pseudo bounding boxes, providing coarse and imperfect spatial labels.}

\trinh{The largest performance advantages of MiTHras over competing models emerge under linear probing, whereas detection results were comparable due to the recall constraint of the shared first-stage candidate detector. Moreover, detection performance universally declined on TCGA-MF-Test compared to MIDOGpp (e.g., MiTHras F$_1$ dropped from $0.829$ to $0.599$), likely due to varying slide quality and acquisition conditions.} Similar degradation has been reported by Jahanifar \textit{et al}. \cite{jahanifar2024mitosis} and Aubreville \textit{et al}. \cite{aubreville2023comprehensive} on multi-domain datasets such as MIDOG21 and MIDOG22. \trinh{This consistent decline across architectures and pretraining strategies indicates a persistent transfer challenge and motivates broader external validation.}

This study has several limitations. 
First, \trinh{the benefits of MiTHras vary across tasks and datasets. Notably, MF subtype classification remains challenging and requires extended validation.}
\trinh{Second, token supervision relies on fixed $50 \times 50$ pseudo boxes, which may include background or exclude parts of larger MFs, thereby introducing supervision noise. Developing more precise spatial labels represents a promising direction for improving token-level supervision.}
\trinh{Third, the audit identified residual errors in the pseudo labels. 55 of 280 reviewed pseudo-mimicker crops contained true MFs, corresponding to $19.6\%$ contamination (95\% Wilson CI, $15.4$--$24.7\%$) and a pool-weighted estimate of $21.7\%$. Per-cohort estimates ranged from $5\%$ to $35\%$, with 20 crops per cohort. These errors can incorrectly group mitotic examples with non-mitotic examples during contrastive learning.}
\trinh{Fourth, although label-interchange experiment demonstrated robustness to pseudo-label noise, this global perturbation may not fully reflect real-world conditions. Label noise can be non-uniform, concentrating on specific organs or challenging morphologies. We leave the development of precise pseudo-label generation and refinement strategies to future investigation.}
Fifth, the main survival comparisons use mitotic count as a single morphological predictor. Although an age- and sex-adjusted analysis assesses its incremental contribution, broader CPath survival models can combine morphology, clinical variables, and transcriptomic profiles \cite{song2024multimodal}. We plan to integrate MiTHras-derived counts into such models to assess their complementary prognostic value.
Finally, \trinh{the clinical utility of MiTHras has yet to be validated; further studies are needed to assess its impact on real-world diagnostic workflows and outcomes.}

\section{Conclusions} \label{section:conclusion}
\trinh{MiTHras combines pseudo-label-guided image- and token-level SCL with masked reconstruction for MF analysis. Pretraining on the 1.8-million-image TCGA-MF-Pseudo corpus, it yields strong performance on external benchmarks, outperforming general-purpose and pathology foundation encoders most significantly under linear probing. Token-level supervision improves some subtype results, while its effect varies across tasks and configurations. Future work will explore finer local labels, noise-robust pseudo-labeling, and broader clinical validation to establish the practical utility of MiTHras.}

\section*{Acknowledgments}
We acknowledge Navid Alemi for his contributions to this project.

\bibliographystyle{IEEEtran}
\bibliography{ms}

\clearpage
\onecolumn

\section*{Supplementary Material}
\setcounter{section}{0}
\setcounter{subsection}{0}
\setcounter{figure}{0}
\setcounter{table}{0}
\setcounter{equation}{0}

\renewcommand{\thesection}{S\arabic{section}}
\renewcommand{\thesubsection}{S\arabic{section}.\arabic{subsection}}
\renewcommand{\thefigure}{S\arabic{figure}}
\renewcommand{\thetable}{S\arabic{table}}
\renewcommand{\theequation}{S\arabic{equation}}
\renewcommand{\theHsection}{supp.\arabic{section}}
\renewcommand{\theHtable}{supp.\arabic{table}}
\renewcommand{\theHfigure}{supp.\arabic{figure}}
\renewcommand{\theHequation}{supp.\arabic{equation}}

\section{Supplementary Results}
\label{sec:supp_results}
Full metric results complement the F$_1$ comparisons in the main manuscript. Metric definitions are provided in Section~\ref{sec:evaluation-metrics}.

\begin{table}[!ht]
\centering
\caption{Fine-tuning and linear-probing results for mitotic figure classification. Values are mean $\pm$ sample SD over three folds. ACC denotes accuracy; ROC-AUC denotes area under the receiver operating characteristic curve. Binary F$_1$, precision, and recall treat mitotic figures as the positive class. Bold indicates the highest mean at displayed precision within each dataset and metric.}
\label{table:results_finetune_full}
\scriptsize
\setlength{\tabcolsep}{4pt}
\renewcommand{\arraystretch}{0.90}
\setlength{\smallskipamount}{1pt}

\begin{adjustbox}{max width=\textwidth}
\begin{tabular}{l|l|cccc|cccc|cccc}
\toprule
\textbf{Fine-tuning} & & \multicolumn{4}{c|}{\textbf{MIDOGpp}} & \multicolumn{4}{c|}{\textbf{Canine}} & \multicolumn{4}{c}{\textbf{TCGA-MF-Test}}\\
Method & Network & ACC & ROC-AUC & Precision & Recall  & ACC & ROC-AUC & Precision & Recall  & ACC & ROC-AUC & Precision & Recall \\
\midrule
\multicolumn{14}{l}{\textit{Natural-image pretraining}} \\
\noalign{\smallskip}
MDFS & EfficientNet-b7 & 0.862$\pm$0.001 & 0.839$\pm$0.112 & 0.865$\pm$0.009 & 0.800$\pm$0.012 & 0.750$\pm$0.007 & 0.776$\pm$0.036 & 0.851$\pm$0.016 & 0.782$\pm$0.032 & 0.811$\pm$0.006 & 0.781$\pm$0.115 & 0.837$\pm$0.015 & 0.691$\pm$0.021 \\
ViT-ImageNet (Supervised) & ViT-base & 0.853$\pm$0.009 & 0.906$\pm$0.036 & 0.838$\pm$0.009 & 0.811$\pm$0.013 & 0.771$\pm$0.006 & 0.818$\pm$0.017 & 0.856$\pm$0.001 & 0.809$\pm$0.011 & 0.817$\pm$0.001 & 0.859$\pm$0.044 & 0.844$\pm$0.017 & 0.702$\pm$0.018 \\
MAE-ImageNet & ViT-base & 0.866$\pm$0.003 & 0.939$\pm$0.005 & 0.851$\pm$0.010 & 0.829$\pm$0.020 & 0.779$\pm$0.008 & 0.841$\pm$0.005 & 0.855$\pm$0.009 & 0.826$\pm$0.025 & 0.819$\pm$0.004 & 0.888$\pm$0.004 & 0.823$\pm$0.002 & 0.733$\pm$0.012 \\
DINOv2 & ViT-base & 0.867$\pm$0.005 & 0.934$\pm$0.005 & 0.858$\pm$0.027 & 0.825$\pm$0.032 & 0.779$\pm$0.007 & 0.828$\pm$0.008 & 0.847$\pm$0.007 & 0.838$\pm$0.021 & 0.824$\pm$0.006 & 0.886$\pm$0.006 & 0.843$\pm$0.013 & 0.723$\pm$0.026 \\
CLIP & ViT-base & 0.856$\pm$0.003 & 0.918$\pm$0.006 & 0.840$\pm$0.006 & 0.818$\pm$0.016 & 0.769$\pm$0.006 & 0.809$\pm$0.016 & 0.842$\pm$0.009 & 0.826$\pm$0.004 & 0.820$\pm$0.005 & 0.863$\pm$0.006 & 0.835$\pm$0.010 & 0.720$\pm$0.001 \\
\midrule
\multicolumn{14}{l}{\textit{Histopathology pretraining models}} \\
\noalign{\smallskip}
CTransPath-MoCov3 & Swin-tiny & 0.841$\pm$0.009 & 0.913$\pm$0.009 & 0.830$\pm$0.009 & 0.788$\pm$0.018 & 0.756$\pm$0.011 & 0.817$\pm$0.008 & 0.843$\pm$0.018 & 0.802$\pm$0.033 & 0.795$\pm$0.006 & 0.858$\pm$0.008 & 0.816$\pm$0.011 & 0.670$\pm$0.007 \\
CHIEF & Swin-tiny & 0.843$\pm$0.011 & 0.918$\pm$0.008 & 0.806$\pm$0.020 & 0.830$\pm$0.014 & 0.755$\pm$0.006 & 0.814$\pm$0.004 & 0.837$\pm$0.010 & 0.809$\pm$0.024 & 0.799$\pm$0.002 & 0.855$\pm$0.009 & 0.837$\pm$0.010 & 0.656$\pm$0.008 \\
Lunit-DINO & ViT-small & 0.846$\pm$0.006 & 0.917$\pm$0.004 & 0.826$\pm$0.015 & 0.807$\pm$0.006 & 0.766$\pm$0.011 & 0.829$\pm$0.007 & 0.854$\pm$0.022 & 0.807$\pm$0.050 & 0.810$\pm$0.004 & 0.861$\pm$0.013 & 0.812$\pm$0.013 & 0.721$\pm$0.026 \\
PLIP & ViT-base/32 & 0.847$\pm$0.011 & 0.920$\pm$0.008 & 0.826$\pm$0.029 & 0.812$\pm$0.012 & 0.755$\pm$0.006 & 0.819$\pm$0.002 & 0.848$\pm$0.009 & 0.793$\pm$0.022 & 0.804$\pm$0.005 & 0.867$\pm$0.008 & 0.819$\pm$0.035 & 0.697$\pm$0.035 \\
MAE-TCGA-Unlabeled & ViT-base & 0.834$\pm$0.016 & 0.912$\pm$0.015 & 0.803$\pm$0.028 & 0.808$\pm$0.012 & 0.758$\pm$0.007 & 0.816$\pm$0.001 & 0.835$\pm$0.013 & 0.819$\pm$0.030 & 0.797$\pm$0.004 & 0.861$\pm$0.006 & 0.791$\pm$0.016 & 0.712$\pm$0.018 \\
SupMAE & ViT-base & 0.793$\pm$0.013 & 0.866$\pm$0.018 & 0.760$\pm$0.019 & 0.751$\pm$0.018 & 0.735$\pm$0.023 & 0.802$\pm$0.009 & 0.849$\pm$0.014 & 0.759$\pm$0.057 & 0.768$\pm$0.014 & 0.849$\pm$0.019 & 0.845$\pm$0.022 & 0.557$\pm$0.041 \\
UNI-DINOv2 & ViT-large & 0.865$\pm$0.010 & 0.918$\pm$0.021 & 0.861$\pm$0.030 & 0.815$\pm$0.013 & 0.775$\pm$0.010 & 0.835$\pm$0.016 & 0.858$\pm$0.015 & 0.814$\pm$0.024 & 0.825$\pm$0.003 & 0.898$\pm$0.010 & \textbf{0.854$\pm$0.018} & 0.712$\pm$0.024 \\
GPFM & ViT-large & 0.828$\pm$0.010 & 0.878$\pm$0.011 & 0.804$\pm$0.007 & 0.785$\pm$0.033 & 0.768$\pm$0.002 & 0.782$\pm$0.009 & 0.833$\pm$0.005 & 0.837$\pm$0.011 & 0.802$\pm$0.008 & 0.841$\pm$0.006 & 0.811$\pm$0.019 & 0.700$\pm$0.002 \\
Virchow & ViT-huge & 0.872$\pm$0.003 & 0.937$\pm$0.001 & 0.855$\pm$0.001 & 0.841$\pm$0.006 & 0.779$\pm$0.005 & 0.831$\pm$0.001 & 0.839$\pm$0.007 & 0.848$\pm$0.018 & 0.830$\pm$0.003 & 0.899$\pm$0.005 & 0.844$\pm$0.014 & 0.737$\pm$0.010 \\
\midrule
\multicolumn{14}{l}{\textit{Pretraining methods using our pseudo-labeled dataset: TCGA-MF-Pseudo}} \\
\noalign{\smallskip}
ViT-TCGA-MF-Pseudo & ViT-base & 0.857$\pm$0.003 & 0.930$\pm$0.002 & 0.861$\pm$0.002 & 0.791$\pm$0.010 & 0.763$\pm$0.005 & 0.834$\pm$0.003 & \textbf{0.867$\pm$0.005} & 0.783$\pm$0.015 & 0.818$\pm$0.002 & 0.888$\pm$0.003 & 0.845$\pm$0.004 & 0.700$\pm$0.008 \\
MAE-TCGA-MF-Pseudo & ViT-base & 0.853$\pm$0.006 & 0.927$\pm$0.005 & 0.824$\pm$0.012 & 0.830$\pm$0.010 & 0.776$\pm$0.004 & 0.834$\pm$0.003 & 0.843$\pm$0.007 & 0.837$\pm$0.015 & 0.805$\pm$0.004 & 0.880$\pm$0.006 & 0.841$\pm$0.002 & 0.670$\pm$0.009 \\
CMAE & ViT-base & 0.863$\pm$0.005 & 0.935$\pm$0.004 & 0.841$\pm$0.011 & 0.835$\pm$0.011 & 0.764$\pm$0.007 & 0.834$\pm$0.002 & 0.863$\pm$0.008 & 0.789$\pm$0.021 & 0.821$\pm$0.004 & 0.887$\pm$0.003 & 0.849$\pm$0.009 & 0.705$\pm$0.022 \\
MiTHras (Ours) & ViT-base & \textbf{0.887$\pm$0.004} & \textbf{0.954$\pm$0.001} & \textbf{0.886$\pm$0.022} & \textbf{0.844$\pm$0.023} & \textbf{0.796$\pm$0.003} & \textbf{0.857$\pm$0.005} & 0.852$\pm$0.005 & \textbf{0.858$\pm$0.003} & \textbf{0.832$\pm$0.004} & \textbf{0.902$\pm$0.000} & 0.848$\pm$0.015 & \textbf{0.739$\pm$0.030} \\
\end{tabular}
\end{adjustbox}
\begin{adjustbox}{max width=\textwidth}
\begin{tabular}{l|l|cccc|cccc|cccc}
\toprule
\textbf{Linear-probing} & & \multicolumn{4}{c|}{\textbf{MIDOGpp}} & \multicolumn{4}{c|}{\textbf{Canine}} & \multicolumn{4}{c}{\textbf{TCGA-MF-Test}}\\
Method & Network & ACC & ROC-AUC & Precision & Recall & ACC & ROC-AUC & Precision & Recall & ACC & ROC-AUC & Precision & Recall \\
\midrule
\multicolumn{14}{l}{\textit{Natural-image pretraining}} \\
\noalign{\smallskip}
MDFS & EfficientNet-b7 & 0.627$\pm$0.045 & 0.615$\pm$0.100 & 0.414$\pm$0.359 & 0.313$\pm$0.274 & 0.618$\pm$0.074 & 0.577$\pm$0.067 & 0.749$\pm$0.041 & 0.708$\pm$0.255 & 0.543$\pm$0.102 &  0.589$\pm$0.102 & 0.496$\pm$0.061 & 0.723$\pm$0.242 \\
ViT-ImageNet (Supervised) & ViT-base & 0.480$\pm$0.082 & 0.490$\pm$0.071 & 0.368$\pm$0.071 & 0.505$\pm$0.446 & 0.500$\pm$0.177 & 0.486$\pm$0.018 & 0.678$\pm$0.022 & 0.524$\pm$0.430 & 0.555$\pm$0.027 & 0.525$\pm$0.042 & 0.429$\pm$0.071 & 0.099$\pm$0.046 \\
MAE-ImageNet & ViT-base & 0.605$\pm$0.007 & 0.519$\pm$0.012 & 0.598$\pm$0.014 & 0.221$\pm$0.076 & 0.535$\pm$0.038 & 0.537$\pm$0.005 & 0.732$\pm$0.004 & 0.533$\pm$0.086 & 0.583$\pm$0.006 & 0.556$\pm$0.019 & 0.518$\pm$0.011 & 0.296$\pm$0.038 \\
DINOv2 & ViT-base & 0.569$\pm$0.007 & 0.533$\pm$0.027 & 0.275$\pm$0.128 & 0.014$\pm$0.013 & 0.562$\pm$0.229 & 0.530$\pm$0.019 & 0.691$\pm$0.021 & 0.653$\pm$0.565 & 0.480$\pm$0.054 & 0.497$\pm$0.057 & 0.415$\pm$0.016 & 0.554$\pm$0.377 \\
CLIP & ViT-base & 0.522$\pm$0.077 & 0.481$\pm$0.054 & 0.385$\pm$0.043 & 0.345$\pm$0.545 & 0.479$\pm$0.203 & 0.487$\pm$0.042 & 0.726$\pm$0.065 & 0.462$\pm$0.500 & 0.487$\pm$0.072 & 0.499$\pm$0.088 & 0.440$\pm$0.045 & 0.640$\pm$0.150 \\
\midrule
\multicolumn{14}{l}{\textit{Histopathology pretraining models}} \\
\noalign{\smallskip}
CTransPath-MoCov3 & Swin-tiny & 0.526$\pm$0.086 & 0.507$\pm$0.059 & 0.342$\pm$0.309 & 0.331$\pm$0.572 & 0.514$\pm$0.037 & 0.547$\pm$0.009 & 0.704$\pm$0.007 & 0.532$\pm$0.097 & 0.432$\pm$0.004 & 0.493$\pm$0.106 & 0.414$\pm$0.019 & \textbf{0.831$\pm$0.215} \\
CHIEF & Swin-tiny & 0.529$\pm$0.091 & 0.506$\pm$0.053 & 0.489$\pm$0.059 & 0.427$\pm$0.457 & 0.523$\pm$0.201 & 0.497$\pm$0.040 & 0.702$\pm$0.015 & 0.569$\pm$0.500 & 0.521$\pm$0.084 & 0.490$\pm$0.043 & 0.463$\pm$0.060 & 0.435$\pm$0.486 \\
Lunit-DINO & ViT-small & 0.511$\pm$0.044 &  0.492$\pm$0.061 & 0.368$\pm$0.103 & 0.436$\pm$0.443 & 0.450$\pm$0.219 & 0.526$\pm$0.064 & 0.684$\pm$0.095 & 0.376$\pm$0.541 & 0.477$\pm$0.017 & 0.553$\pm$0.069 & 0.411$\pm$0.049 & 0.637$\pm$0.330 \\
PLIP & ViT-base/32 & 0.494$\pm$0.075 & 0.538$\pm$0.031 & 0.420$\pm$0.022 & 0.598$\pm$0.522 & 0.487$\pm$0.202 & 0.511$\pm$0.063 & 0.654$\pm$0.052 & 0.491$\pm$0.494 & 0.457$\pm$0.096 & 0.456$\pm$0.063 & 0.433$\pm$0.043 & 0.670$\pm$0.406 \\
MAE-TCGA-Unlabeled & ViT-base & 0.533$\pm$0.017 & 0.505$\pm$0.011 & 0.438$\pm$0.009 & 0.343$\pm$0.111 & 0.596$\pm$0.046 & 0.613$\pm$0.004 & 0.729$\pm$0.008 & 0.677$\pm$0.118 & 0.611$\pm$0.008 & 0.646$\pm$0.011 & 0.570$\pm$0.041 & 0.411$\pm$0.131  \\
SupMAE & ViT-base & 0.654$\pm$0.010 & 0.715$\pm$0.009 & 0.579$\pm$0.013 & 0.677$\pm$0.023 & 0.667$\pm$0.008 & 0.739$\pm$0.006 & 0.833$\pm$0.010 & 0.658$\pm$0.025 & 0.661$\pm$0.005 & 0.677$\pm$0.002 & 0.609$\pm$0.016 & 0.577$\pm$0.039  \\
UNI-DINOv2 & ViT-large & 0.511$\pm$0.076 & 0.577$\pm$0.021 & 0.454$\pm$0.067 & 0.455$\pm$0.476 & 0.586$\pm$0.194 & 0.486$\pm$0.007 & 0.719$\pm$0.030 & 0.705$\pm$0.494 & 0.484$\pm$0.077 & 0.501$\pm$0.051 & 0.428$\pm$0.028 & 0.531$\pm$0.277 \\
GPFM & ViT-large & 0.471$\pm$0.076 & 0.495$\pm$0.028 & 0.414$\pm$0.014 & 0.573$\pm$0.403 & 0.593$\pm$0.100 & 0.492$\pm$0.008 & 0.702$\pm$0.006 & 0.727$\pm$0.238 & 0.538$\pm$0.063 & 0.489$\pm$0.114 & 0.464$\pm$0.083 & 0.355$\pm$0.090 \\
Virchow & ViT-huge & 0.532$\pm$0.056 & 0.536$\pm$0.008 & 0.446$\pm$0.015 & 0.339$\pm$0.388 & 0.494$\pm$0.176 & 0.502$\pm$0.030 & 0.686$\pm$0.024 & 0.489$\pm$0.441 & 0.527$\pm$0.090 & 0.562$\pm$0.014 & 0.524$\pm$0.116 & 0.365$\pm$0.534 \\
\midrule
\multicolumn{14}{l}{\textit{Pretraining methods using our pseudo-labeled dataset: TCGA-MF-Pseudo}} \\
\noalign{\smallskip}
ViT-TCGA-MF-Pseudo & ViT-base & 0.854$\pm$0.001 & \textbf{0.925$\pm$0.001} & 0.842$\pm$0.009 & \textbf{0.807$\pm$0.011} & 0.758$\pm$0.004 & 0.817$\pm$0.001 & 0.853$\pm$0.004 & 0.792$\pm$0.013 & 0.809$\pm$0.002 & 0.870$\pm$0.002 & \textbf{0.802$\pm$0.003} & 0.734$\pm$0.007 \\
MAE-TCGA-MF-Pseudo & ViT-base & 0.554$\pm$0.014 & 0.581$\pm$0.016 & 0.457$\pm$0.016 & 0.263$\pm$0.113 & 0.635$\pm$0.038 & 0.591$\pm$0.009 & 0.731$\pm$0.007 & 0.761$\pm$0.101 & 0.540$\pm$0.003 & 0.650$\pm$0.012 & 0.471$\pm$0.003 & 0.657$\pm$0.058 \\
CMAE & ViT-base & 0.605$\pm$0.002 & 0.560$\pm$0.008 & 0.550$\pm$0.010 & 0.391$\pm$0.094  & 0.558$\pm$0.023 & 0.638$\pm$0.007 & 0.789$\pm$0.014 & 0.506$\pm$0.053 & 0.678$\pm$0.004 & 0.714$\pm$0.008 & 0.691$\pm$0.063 & 0.465$\pm$0.114 \\
MiTHras (Ours) & ViT-base & \textbf{0.866$\pm$0.002} & \textbf{0.925$\pm$0.001} & \textbf{0.872$\pm$0.005} & 0.802$\pm$0.010 & \textbf{0.788$\pm$0.005} & \textbf{0.845$\pm$0.001} & \textbf{0.856$\pm$0.006} & \textbf{0.840$\pm$0.016} & \textbf{0.814$\pm$0.002} & \textbf{0.878$\pm$0.001} & 0.800$\pm$0.006 & 0.752$\pm$0.007  \\
\bottomrule
\end{tabular}
\end{adjustbox}

\end{table}

\begin{table}[!ht]
\begin{center}
\caption{Mitotic figure detection and subtype classification results, mean $\pm$ standard deviation over three folds. The segmentation-only baseline matches the configuration reported in the main detection table. For two-stage models, the detection/classification threshold pair maximizes F$_1$ across the three MIDOG22 validation folds and is held fixed for external evaluation; YOLOv10 uses its own inference confidence threshold. SupMAE is pretrained on labeled MIDOG22. Bold indicates the highest mean at displayed precision in each column. AMi-Br uses atypical mitotic figures as the positive class (label 1), with typical figures labeled 0. AMi-Morph uses unweighted macro F$_1$, precision, and recall; ROC-AUC was not evaluated. ACC denotes overall accuracy.}
\label{table:results_mitosis_detection_subtype}
\setlength{\tabcolsep}{2pt}
\renewcommand{\arraystretch}{1}
\begin{adjustbox}{width=\textwidth}
\begin{tabular}{l|c|cc|cc|cc|cccc|ccc}
\toprule
\multirow{2}{*}{Pretraining Method} & \multirow{2}{*}{Network} & \multicolumn{2}{c|}{MIDOGpp} & \multicolumn{2}{c|}{Canine} & \multicolumn{2}{c|}{TCGA-MF-Test} & \multicolumn{4}{c|}{AMi-Br-TUPAC} & \multicolumn{3}{c}{AMi-Morph-TUPAC} \\
& &Recall & Precision & Recall & Precision & Recall & Precision & ACC & ROC-AUC & Precision & Recall & ACC & Precision & Recall \\
\midrule
\multicolumn{8}{l}{\textit{Detection baselines}} \\
Segmentation only & Efficient-UNet-b0 & 0.715$\pm$0.000 & \textbf{0.864$\pm$0.000} & 0.789$\pm$0.000 & 0.776$\pm$0.000 & 0.620$\pm$0.000 & 0.529$\pm$0.000 & -& -& -& -& -& -& - \\
YOLOv10 & YOLOv10 & 0.801$\pm$0.061 & 0.738$\pm$0.100 & 0.812$\pm$0.000 & 0.772$\pm$0.000 & 0.704$\pm$0.000 & 0.468$\pm$0.000  & -& -& -& -& -& -& -\\
\midrule
\multicolumn{8}{l}{\textit{Natural-image pretraining}} \\
Bertram et al.\ \cite{bertram2025histologic} & EfficientNet-V2-S & - & - & - & - & - & - & 0.761$\pm$0.028 & 0.840$\pm$0.011 & 0.468$\pm$0.034 & \textbf{0.758$\pm$0.041} & 0.536$\pm$0.040 & 0.311$\pm$0.016 & 0.348$\pm$0.037 \\
MDFS & EfficientNet-b7 & 0.778$\pm$0.001 & 0.846$\pm$0.006 & 0.795$\pm$0.019 & 0.776$\pm$0.013 & 0.679$\pm$0.012 & 0.523$\pm$0.012 & 0.815$\pm$0.003 & 0.838$\pm$0.012 & 0.565$\pm$0.019 & 0.611$\pm$0.098 & 0.692$\pm$0.036 & 0.452$\pm$0.035 & 0.422$\pm$0.030 \\
ViT-ImageNet (Supervised) & ViT-base & 0.783$\pm$0.022 & 0.842$\pm$0.022 & 0.815$\pm$0.034 & 0.766$\pm$0.035 & 0.689$\pm$0.035 & 0.506$\pm$0.022 & 0.834$\pm$0.011 & 0.858$\pm$0.011 & 0.609$\pm$0.043 & 0.643$\pm$0.053 & 0.707$\pm$0.026 & 0.502$\pm$0.018 & 0.460$\pm$0.025 \\
MAE-ImageNet & ViT-base & 0.777$\pm$0.014 & 0.847$\pm$0.004 & 0.799$\pm$0.018 & 0.782$\pm$0.004 & \textbf{0.705$\pm$0.007} & 0.499$\pm$0.004 & 0.820$\pm$0.043 & 0.784$\pm$0.153 & 0.572$\pm$0.169 & 0.400$\pm$0.253 & 0.692$\pm$0.007 & 0.438$\pm$0.060 & 0.451$\pm$0.023  \\
DINOv2 & ViT-base & 0.793$\pm$0.016 & 0.835$\pm$0.015 &  0.832$\pm$0.010 & 0.762$\pm$0.006 & \textbf{0.705$\pm$0.010} & 0.510$\pm$0.006 & 0.855$\pm$0.007 & 0.887$\pm$0.003 & 0.669$\pm$0.034 & 0.643$\pm$0.039 & 0.684$\pm$0.063 & 0.498$\pm$0.041 & 0.486$\pm$0.030 \\
CLIP & ViT-base & 0.758$\pm$0.007 & 0.863$\pm$0.001 & 0.796$\pm$0.005 & 0.786$\pm$0.007 & 0.666$\pm$0.002 & 0.529$\pm$0.003 & 0.850$\pm$0.006 & 0.886$\pm$0.011 & 0.656$\pm$0.019 & 0.631$\pm$0.038 & \textbf{0.719$\pm$0.020} & 0.497$\pm$0.011 & 0.460$\pm$0.004 \\
\midrule
\multicolumn{8}{l}{\textit{Histopathology pretraining models}} \\
CTransPath-MoCov3 & Swin-tiny & 0.732$\pm$0.011 & 0.855$\pm$0.006 & 0.786$\pm$0.012 & 0.796$\pm$0.008 & 0.635$\pm$0.002 & 0.525$\pm$0.003 & 0.801$\pm$0.013 & 0.685$\pm$0.128 & 0.433$\pm$0.377 & 0.157$\pm$0.145 & 0.663$\pm$0.023 & 0.309$\pm$0.048 & 0.375$\pm$0.052  \\
CHIEF & Swin-tiny & 0.744$\pm$0.006 & 0.859$\pm$0.005 & 0.785$\pm$0.014 & 0.797$\pm$0.007 & 0.622$\pm$0.006 & 0.529$\pm$0.006 & 0.810$\pm$0.009 & 0.811$\pm$0.005 & 0.563$\pm$0.031 & 0.526$\pm$0.040 & 0.644$\pm$0.021 &  0.345$\pm$0.040 & 0.315$\pm$0.021 \\
Lunit-DINO & ViT-small & 0.766$\pm$0.009 & 0.842$\pm$0.005 & 0.795$\pm$0.029 & 0.784$\pm$0.017 & 0.671$\pm$0.014 & 0.502$\pm$0.002 & 0.796$\pm$0.013 & 0.833$\pm$0.009 & 0.521$\pm$0.028 & 0.644$\pm$0.045 & 0.663$\pm$0.008 & 0.389$\pm$0.016 & 0.378$\pm$0.034 \\
PLIP & ViT-base/32 & 0.765$\pm$0.004 & 0.843$\pm$0.011 & 0.786$\pm$0.014 & 0.782$\pm$0.002  & 0.669$\pm$0.018 & 0.508$\pm$0.015 & 0.850$\pm$0.009  & 0.882$\pm$0.005 & 0.659$\pm$0.044 & 0.630$\pm$0.040 & 0.703$\pm$0.029 & 0.470$\pm$0.026 & 0.456$\pm$0.024 \\
MAE-TCGA-Unlabeled & ViT-base & 0.790$\pm$0.005 & 0.809$\pm$0.008 & 0.839$\pm$0.014 & 0.746$\pm$0.011 & 0.701$\pm$0.008 & 0.477$\pm$0.007 & 0.788$\pm$0.017 & 0.724$\pm$0.039 & 0.545$\pm$0.117 & 0.164$\pm$0.051 & 0.664$\pm$0.011 & 0.310$\pm$0.006 & 0.337$\pm$0.006 \\
SupMAE & ViT-base & 0.702$\pm$0.008 & 0.846$\pm$0.006 & 0.750$\pm$0.032 & \textbf{0.801$\pm$0.011} & 0.551$\pm$0.021 & \textbf{0.530$\pm$0.009} & 0.784$\pm$0.017 & 0.793$\pm$0.008 & 0.498$\pm$0.033 & 0.520$\pm$0.084 & 0.681$\pm$0.011 & 0.388$\pm$0.023 & 0.418$\pm$0.032 \\
UNI-DINOv2 & ViT-large & 0.787$\pm$0.007 & 0.843$\pm$0.012 & 0.807$\pm$0.013 & 0.783$\pm$0.014  & 0.692$\pm$0.012 & 0.510$\pm$0.010 & 0.810$\pm$0.003 & 0.824$\pm$0.008 & 0.554$\pm$0.011 & 0.586$\pm$0.061 & 0.654$\pm$0.009 & 0.386$\pm$0.007 & 0.391$\pm$0.015 \\
GPFM & ViT-large & 0.721$\pm$0.016 & 0.859$\pm$0.002 & 0.803$\pm$0.004 & 0.792$\pm$0.004  & 0.642$\pm$0.004 & 0.520$\pm$0.008 & 0.847$\pm$0.003 &0.890$\pm$0.006 & 0.636$\pm$0.020 & 0.674$\pm$0.046 & 0.701$\pm$0.025 & 0.504$\pm$0.015 & 0.472$\pm$0.030 \\
Virchow & ViT-huge & 0.792$\pm$0.007 & 0.842$\pm$0.006 & \textbf{0.842$\pm$0.009} & 0.755$\pm$0.005  & 0.690$\pm$0.008 & 0.514$\pm$0.003 & 0.781$\pm$0.009 & 0.829$\pm$0.023 & 0.492$\pm$0.015 & 0.678$\pm$0.033 & 0.611$\pm$0.038 & 0.371$\pm$0.022 & 0.345$\pm$0.020 \\
\midrule
\multicolumn{8}{l}{\textit{Pretraining methods using our pseudo-labeled dataset: TCGA-MF-Pseudo}} \\
ViT-TCGA-MF-Pseudo & ViT-base & 0.761$\pm$0.005 & 0.863$\pm$0.001 &  0.793$\pm$0.005 & 0.789$\pm$0.002 & 0.669$\pm$0.007 & 0.522$\pm$0.002 & 0.806$\pm$0.008 & 0.828$\pm$0.001 & 0.555$\pm$0.037 & 0.527$\pm$0.092 & 0.627$\pm$0.018 & 0.262$\pm$0.006 & 0.330$\pm$0.010 \\
MAE-TCGA-MF-Pseudo & ViT-base & 0.803$\pm$0.007 & 0.804$\pm$0.006 & 0.838$\pm$0.010 & 0.750$\pm$0.005 & 0.668$\pm$0.006 & 0.496$\pm$0.003 & 0.787$\pm$0.001 & 0.637$\pm$0.058 & 0.527$\pm$0.016 & 0.057$\pm$0.035 & 0.580$\pm$0.109 & 0.218$\pm$0.091 & 0.260$\pm$0.101 \\
CMAE & ViT-base & \textbf{0.810$\pm$0.003} & 0.810$\pm$0.006 & 0.797$\pm$0.013 & 0.774$\pm$0.005 & 0.699$\pm$0.009 & 0.498$\pm$0.013 & 0.832$\pm$0.017 & 0.857$\pm$0.008 & 0.620$\pm$0.059 & 0.583$\pm$0.062 & 0.697$\pm$0.010 & 0.346$\pm$0.014 & 0.407$\pm$0.008  \\
MiTHras (Ours) & ViT-base & 0.798$\pm$0.019 & \textbf{0.864$\pm$0.018} & 0.810$\pm$0.010 & 0.786$\pm$0.010 & 0.692$\pm$0.023 & 0.529$\pm$0.017  & \textbf{0.870$\pm$0.011} & \textbf{0.914$\pm$0.004} & \textbf{0.696$\pm$0.052} & 0.710$\pm$0.062 & 0.713$\pm$0.009 & \textbf{0.511$\pm$0.006} & \textbf{0.509$\pm$0.020} \\
\bottomrule
\end{tabular}
\end{adjustbox}
\end{center}
\end{table}

\end{document}